\documentclass[11pt]{article}

\usepackage[final]{acl}

\usepackage{times}
\usepackage{latexsym}

\usepackage[T1]{fontenc}

\usepackage[utf8]{inputenc}

\usepackage{microtype}

\usepackage{inconsolata}

\usepackage{graphicx}
\usepackage{subcaption}
\usepackage{ltablex}
\keepXColumns
\usepackage{algorithm}

\usepackage{algpseudocode}
\usepackage{amsmath}
\title{No Reader Left Behind: Multi-Agent Summaries Everyone Can Understand}

\author{
Jimin Jung$^{1}$, MyoungJin Kim$^{1}$, Jaehyung Seo$^{2\dagger}$, Heuiseok Lim$^{1\dagger}$ \\
$^{1}$Department of Computer Science and Engineering, Korea University \\
$^{2}$Department of Computer Science and Engineering, Konkuk University \\
\texttt{\{stopmin02, imannamj, limhseok\}@korea.ac.kr} \\
\texttt{seojae777@konkuk.ac.kr}
}

\usepackage{booktabs}
\usepackage{multirow}

\begin{document}
\maketitle
\begingroup
\renewcommand{\thefootnote}{}
\footnotetext{\textdagger\ Corresponding Authors}
\endgroup

\begin{abstract}
The Plain Writing Act in the United States requires government documents to be written in clear and simple language. However, existing summarization systems struggle to address diverse linguistic and cognitive barriers among general readers. We propose NRLB (No Reader Left Behind), a unified multi-agent framework for plain language summarization that simulates three representative reader groups: elementary school students, non-native speakers, and readers with attention deficits. NRLB integrates template-based planning with an iterative feedback loop guided by simulated readers and domain expert revision to address comprehension barriers such as unknown terms, missing contexts, and confusing sentences. Evaluations across multiple datasets demonstrate consistent improvements in both readability and factuality. Human evaluation further supports these findings, with annotator preference rates ranging from 55\% to 76\%, highlighting NRLB’s ability to generate summaries that are both faithful to the source and accessible to a wide range of readers.
\end{abstract}

\begin{figure}[t]
    \centering
    \includegraphics[width=0.85\columnwidth]{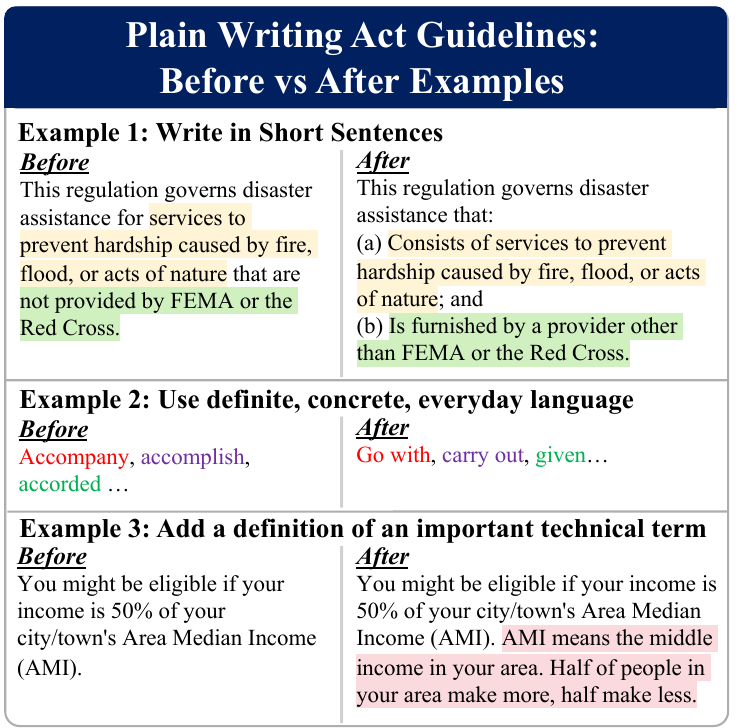}
    \caption{Examples of revision strategies based on the U.S. Plain Writing Act, including shortening sentences, using concrete language, and defining technical terms.}
    \label{fig:intro}
\end{figure}

\section{Introduction}
During the COVID-19 pandemic, gaps in scientific communication deepened public confusion. Lower science literacy was associated with greater susceptibility to misinformation and lower compliance, whereas even a 1\% increase in public science literacy led to a 14.2\% rise in civic engagement with public health measures \cite{qin2024mass}. These findings suggest that even modest improvements in science literacy can directly influence public behavior, underscoring the importance of clear and accessible scientific communication for effective public health responses \cite{xu2015problems}. The Plain Writing Act, enacted by the U.S. government in 2010, mandates that official communications be understandable to the general public. These requirements reflect a long-standing institutional need for accessible public communication. Federal guidelines recommend a 6th-8th grade reading level, concise and active sentences, simplified or annotated technical terms, and accessibility across platforms. Agencies are also required to provide plain language resources and mechanisms for public feedback.\footnote{\url{https://www.justice.gov/open/plain-writing-act}} These guidelines emphasize the need to accommodate readers with limited literacy, non-native English proficiency, or cognitive and visual impairments \cite{gooding2022ethical}. These principles are typically implemented through strategies such as sentence simplification, vocabulary reduction, and defining specialized terms, as illustrated in Figure~\ref{fig:intro}.

However, existing automated summarization and simplification systems often fail to adhere to the principles of the Plain Writing Act. They often overlook key reader-specific factors such as background knowledge, attention span, and syntactic complexity, resulting in outputs that retain technical jargon and complex sentence structures. While some models achieve surface-level simplification, they rarely improve comprehension for general audiences. Furthermore, many prior approaches generate multiple versions tailored to specific subgroups (e.g., K-12 students or non-native adults), which conflicts with the Act’s core principle of clarity for all readers \cite{mo2024expertease, xu2015problems}. These limitations highlight the need for a unified framework that supports broader accessibility through adaptive simplification.

To address this gap, we propose NRLB (No Reader Left Behind), an automated framework for plain language summarization that explicitly models commonly overlooked reading barriers. Grounded in prior research on literacy and cognitive processing \cite{smith2021role,tighe2016examining,shero2021differential}, NRLB simulates three representative reader groups: elementary school students, non-native readers, and readers with attention deficits. Each agent identifies distinct comprehension challenges, enabling targeted revisions that improve accessibility across diverse reader populations.

The NRLB framework consists of two core modules. \textbf{Module 1: Content Planning and Drafting} begins with a Planner Agent that classifies the input document and selects an appropriate Domain Expert Agent, who generates a template-based initial summary capturing the core content. \textbf{Module 2: Feedback-Guided Simplification} refines this draft through a structured loop of reader feedback and expert revision. Reader Agents identify segments with lexical complexity, missing context, or syntactic difficulty. A Checklist Agent aggregates and prioritizes these issues, which are then used by the Domain Expert Agent to propose targeted edits. Finally, an Editor Agent applies these revisions in context to produce the final plain language summary, which is evaluated using metrics for readability, relevance, and factuality. This unified design enables systematic identification and resolution of diverse comprehension barriers across reader groups.

Our contributions are as follows:

(1) We propose NRLB, a unified plain language summarization framework that simulates diverse reader perspectives within a single system to explicitly model key comprehension barriers across reader groups. 

(2) We introduce a structured multi-agent feedback loop with coordinated role-based agents that enables iterative identification and systematic resolution of comprehension barriers. 

(3) We demonstrate that NRLB improves both readability and factuality through multi-round refinement across multiple benchmarks and human evaluations.

\begin{figure*}[t]
    \centering
    \includegraphics[width=1.0\textwidth]{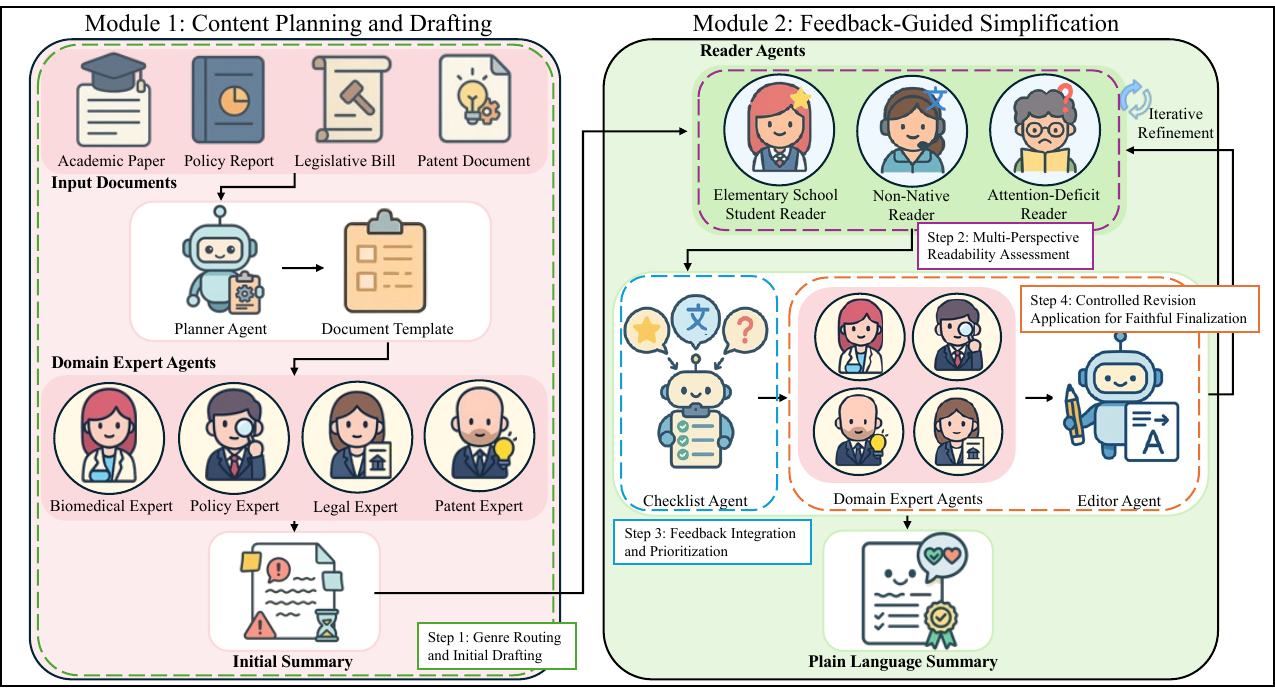}
    \caption{Overview of NRLB. Module 1 produces an initial draft via a Planner and Domain Expert Agent, while Module 2 iteratively refines it through a feedback loop with Reader, Checklist, Domain Expert, and Editor Agents.}
    \label{fig:main}
\end{figure*}

\section{Related Work}
\subsection{Plain Language Summarization}
Plain language summarization aims to simplify language while preserving key content and meaning. Prior work has focused on lexical or syntactic rewriting, often using Simple Wikipedia \cite{xu2015problems}, readability-controlled fine-tuning \cite{luo2022readability}, or task-specific evaluation metrics \cite{urlana2023controllable}. However, these approaches tend to emphasize surface-level readability, often at the expense of content fidelity.

In contrast, NRLB preserves the core content of documents through a template-based initial summary and maintains semantic consistency through a structured, controlled revision process. Rather than relying on simple text reduction, it addresses barriers such as unknown terms, missing context, and complex sentence structures, enabling both readability and factuality.

\subsection{Multi-Agent Summarization}
To address these limitations, recent work has explored multi-agent architectures where role-specific agents collaborate to overcome the limitations of single large language models \cite{du2024multi, chen2023agentverse}. In text simplification, \textbf{AgentSimp} \cite{fang2025collaborative} adopts a stage-wise pipeline with sequential agents (e.g., Metaphor Analysis, Article Logic Analyst), enabling structured multi-agent collaboration through intermediate analysis and rewriting. Although AgentSimp supports iterative reconstruction, it lacks an explicit feedback-driven refinement mechanism that systematically incorporates accumulated feedback across iterations, limiting its ability to progressively improve readability. Similarly, \textbf{ExpertEase} \cite{mo2024expertease} employs teacher, student, and expert roles to generate grade-specific summaries tailored to different educational levels. However, its audience-segregated design diverges from the goal of the Plain Writing Act, which seeks universally understandable documents.

In contrast, \textbf{NRLB} introduces a feedback-driven iterative refinement loop that accumulates and incorporates simulated reader feedback across steps, enabling systematic identification and resolution of comprehension barriers, resulting in more adaptive and inclusive plain language summarization.

\subsection{Modeling Reader Diversity}
Text simplification has traditionally aimed to support underserved readers, such as individuals with dyslexia, non-native speakers, and those with cognitive challenges. For example, \cite{rello2013frequent} showed that replacing unfamiliar words with shorter and more frequent alternatives improved reading speed and comprehension, while \cite{wilkens2020simplifying} demonstrated the effectiveness of vocabulary-based simplification strategies for readers with cognitive challenges in multilingual settings. However, prior work has largely focused on lexical-level simplification and often treats different reader groups in isolation, failing to adequately address broader comprehension barriers such as limited background knowledge or syntactic complexity. Although these challenges have been noted in prior studies \cite{xu2015problems, dadu2021text}, few approaches systematically address them within a unified framework.

In contrast, NRLB jointly models reader groups, including elementary school students, non-native readers, and readers with attention difficulties within a unified framework. It simulates feedback from Reader Agents to systematically identify and resolve comprehension barriers. While collecting human feedback iteratively is costly and often impractical, recent work suggests that LLMs can approximate human reading behavior when guided by persona attributes \cite{argyle2023out, park2023generative, aher2023using}, enabling feedback-driven iterative refinement. This approach enables unified modeling and systematic resolution of diverse comprehension barriers.

\section{Methodology: NRLB}
Figure~\ref{fig:main} presents an overview of the NRLB (No Reader Left Behind) framework, which generates plain language summaries for general audiences through two sequential modules. This design enables structured coordination between planning, feedback, and revision, ensuring both content preservation and readability. \textbf{Module 1: Content Planning and Drafting} begins with a Planner Agent that classifies the genre of the input document. Based on this classification, the system selects an appropriate Domain Expert Agent and assigns a predefined template corresponding to the identified document type. The Domain Expert Agent then completes the template by following genre-specific guidelines, producing an initial summary that emphasizes informational completeness and logical flow. \textbf{Module 2: Feedback-Guided Simplification} provides the initial summary to three simulated Reader Agents: Elementary School Student Reader, Non-Native Reader, and Attention-Deficit Reader. These agents independently identify comprehension barriers across three categories: unknown terms, missing contexts, and confusing sentences. The Checklist Agent integrates and prioritizes the feedback based on their frequency and estimated difficulty. The resulting checklist is returned to the Domain Expert Agent, who drafts specific revision suggestions. The Editor Agent then applies these suggestions to produce a revised summary. This process is repeated for up to N iterations, following the draft, review, and revise cycle outlined in the U.S. Plain Writing Act (2010). Additional implementation details are provided in Appendix~\ref{sec:A}.

\subsection{Genre Routing and Initial Drafting}
The NRLB framework begins with a genre routing step that determines the document type and corresponding expert configuration. The Planner Agent classifies the input document into one of four types: Academic Paper, Policy Report, Legislative Bill, or Patent Document, and assigns the corresponding Domain Expert Agent (e.g., Biomedical, Policy, Legal, or Patent Expert). For example, a Policy Report is routed to a Policy Expert. Based on the genre label, the Planner Agent selects a slot-based summary template aligned with standard writing conventions for that document type. Each template consists of labeled sections with recommended sentence lengths. For instance, a policy report from the U.S. Government Accountability Office (GAO) typically includes “Why GAO Did This Study,” “What GAO Found,” and “What GAO Recommends.” A list of genre-specific templates is provided in Appendix~\ref{sec:A1}.

The Domain Expert Agent completes the template using example summaries and writing instructions (Appendix~\ref{sec:A2}). The agent aims to ensure informational completeness and logical flow while adhering to length constraints. In biomedical academic papers, the Methods section summarizes the experimental design, while the Results section highlights key findings. Although the initial summary captures the core content, it may still include barriers to comprehension such as unknown terms, missing contexts, or confusing sentences. These issues are addressed through iterative refinement in the next module.

\subsection{Multi-Perspective Readability Assessment}\label{sec:reader_agent}
To capture diverse comprehension barriers faced by general readers, we identified three representative reader profiles in Module 2. These profiles are grounded in prior research on literacy and cognitive processing and include the Elementary School Student Reader Agent, the Non-Native Reader Agent, and the Attention-Deficit Reader Agent, each representing a distinct reader population. All three agents evaluate the same summary according to the linguistic and cognitive characteristics of the group they simulate. They are instantiated using an LLM with prompts that reflect the reader’s abilities at three levels: word-level (e.g., limited vocabulary), knowledge-level (e.g., lack of background context), and sentence-level (e.g., difficulty parsing complex syntax). Each agent is instructed to detect comprehension barriers across all three predefined categories: unknown terms, missing contexts, and confusing sentences. This simulation-based approach enables consistent and reproducible feedback across diverse reader perspectives without requiring human annotators. Full prompt specifications are provided in Appendix~\ref{sec:A3}.

\begin{itemize}
    \item \textbf{The Elementary School Student Reader Agent} simulates the perspective of a fourth-grade student with a vocabulary of approximately 3,000 common English words  \cite{laufer2010lexical}. It flags unfamiliar or technical terms beyond this range as unknown terms. Institutional references or scientific concepts that require background knowledge are classified as missing contexts. Sentences longer than 15 words or containing multiple subordinate clauses are marked as confusing sentences. Typical feedback includes phrases such as “too many difficult words,” “background explanation is needed,” and “the sentence is too long.”
    \item \textbf{The Non-Native Reader Agent} represents an adult reader familiar with high-frequency vocabulary but sensitive to low-frequency academic terms, idioms, and cultural expressions \cite{ha2022vocabulary}. It flags rare or idiomatic phrases as unknown terms and identifies culturally specific institutions or historical references lacking explanation as missing contexts. Sentences with garden-path constructions, reduced relative clauses, or repeated passive voice are labeled as confusing sentences. Common feedback includes statements like “I don’t know this word,” “the cultural context is missing,” and “the sentence structure is too complicated.”
    \item \textbf{The Attention-Deficit Reader Agent} reflects the cognitive profile of individuals with limited working memory and reduced attention span \cite{jacobson2011working}. It marks visually or phonetically complex words, compounds, and low-frequency vocabulary as unknown terms. Concepts introduced without sufficient background are flagged as missing contexts. Sentences exceeding 15 words or containing multiple relative clauses are categorized as confusing sentences. Representative feedback includes comments such as “the word is hard to decode” and “the long sentence makes it hard to stay focused.”
\end{itemize}
Each Reader Agent provides structured feedback, which is integrated and prioritized in the next stage. This consolidated view reveals comprehension barriers that general readers may face. Although we focus on three representative reader types, NRLB is designed to easily plug in additional Reader Agents, allowing flexible extension to additional reader profiles.

\subsection{Feedback Integration and Prioritization}
The Checklist Agent collects structured feedback from the Reader Agents, where issues are categorized as unknown terms, missing contexts, or confusing sentences. The priority of each issue is determined by the number of agents that flagged it. Issues flagged by all three agents receive top priority. When multiple issues have the same priority, the agent applies the Automated Readability Index (ARI) to assess sentence complexity \cite{senter1967automated}. The ARI estimates U.S. grade-level difficulty by combining the average characters per word and the average number of words per sentence. For each category, the Checklist Agent selects up to three items with the highest difficulty and compiles them into a unified checklist. This refined list is forwarded to the Domain Expert Agent to guide targeted revisions.

\subsection{Controlled Revision Application for Faithful Finalization}
The Domain Expert Agent generates revision proposals based on the checklist received from the Checklist Agent. Using the same domain expert as in Module 1 ensures continuity between drafting and revision. It replaces unknown terms with simple synonyms or 5-word definitions, adds missing contexts with up to 15-word explanations from the source, rewrites confusing sentences into shorter active sentences, and marks items as “insufficient information” if no source support exists.

The Editor Agent applies these revisions to produce an updated summary while maintaining sentence-level coherence and contextual flow. When edits involve splitting or replacing parts of the same sentence, the agent determines a conflict-free application order. Revisions are only applied when the original sentence exists in the current summary, and replacements are executed sequentially to preserve consistency. The Editor Agent integrates reader feedback with expert-suggested revisions to enable coherent multi-perspective refinement. This revision process can be repeated up to N times; in our experiments, we used two iterations by default. This process improves readability while preserving factual consistency through controlled integration of revisions. As shown in Section~\ref{sec:ablation}, removing the Editor Agent degrades the factual consistency of the final summary.

\section{Experimental Setup}
\subsection{Datasets}
We evaluate NRLB on four representative datasets across diverse domains. PLOS \cite{goldsack2022making} consists of biomedical academic papers, GovReport \cite{huang2021efficient} consists of policy reports from U.S. government agencies, BillSum \cite{kornilova2019billsum} consists of U.S. legislative bills, and BigPatent \cite{sharma2019bigpatent} consists of patent documents with rich technical language. We randomly sample 500 test examples from each dataset to ensure both statistical robustness and computational efficiency, following recent best practices in long-document summarization research \cite{xu2025evaluating}. All datasets include human-written reference summaries, and further dataset-specific details are provided in Appendix~\ref{sec:C1}.

\begin{table*}[t]
\centering
\scriptsize
\setlength{\tabcolsep}{3pt}
\renewcommand{\arraystretch}{1.05}

\begin{tabular}{ll*{12}{c}}
\toprule
\multicolumn{2}{c}{} 
  & \multicolumn{3}{c}{\textbf{PLOS}}
  & \multicolumn{3}{c}{\textbf{GovReport}}
  & \multicolumn{3}{c}{\textbf{BillSum}}
  & \multicolumn{3}{c}{\textbf{BigPatent}} \\
\cmidrule(lr){3-5} \cmidrule(lr){6-8} \cmidrule(lr){9-11} \cmidrule(lr){12-14}
\multicolumn{2}{c}{} 
  & GPT-4o & Llama-3.1 & Qwen3
  & GPT-4o & Llama-3.1 & Qwen3
  & GPT-4o & Llama-3.1 & Qwen3
  & GPT-4o & Llama-3.1 & Qwen3 \\
\midrule

  & Initial Summary  & \textbf{46.85} & \textbf{47.27} & \textbf{41.62}
           & \textbf{37.93} & \textbf{46.93} & \textbf{46.19}
           & \textbf{46.70} & \textbf{45.81} & \textbf{41.08}
           & \textbf{41.87} & \textbf{44.94} & \textbf{34.21} \\
\textbf{ROUGE-1}
  & Round 1    & 45.12 & 44.43 & 37.55
           & 36.35 & 45.33 & 41.41
           & 43.28 & 42.66 & 37.08
           & 37.74 & 40.09 & 30.55 \\
  & Round 2    
           & 42.77 & 41.64 & 36.03
           & 34.73 & 44.19 & 40.04
           & 40.88 & 39.70 & 34.62
           & 35.40 & 36.87 & 29.18 \\
\midrule

  & Initial Summary  & \textbf{86.91} & \textbf{86.51} & \textbf{85.78}
           & \textbf{86.14} & \textbf{86.30} & \textbf{85.59}
           & \textbf{87.07} & \textbf{86.83} & \textbf{85.61}
           & \textbf{86.52} & \textbf{86.68} & \textbf{85.55} \\
\textbf{BERTScore}           
  & Round 1    & 86.61 & 86.00 & 85.33
           & 85.89 & 85.84 & 85.04
           & 86.43 & 86.20 & 85.03
           & 85.82 & 85.73 & 84.89 \\
  & Round 2
           & 86.29 & 85.43 & 84.96
           & 85.63 & 85.48 & 84.74
           & 85.99 & 85.59 & 84.62
           & 85.44 & 85.11 & 84.50 \\
\midrule

  & Initial Summary  & 17.73 & 18.74 & 19.12
           & 17.79 & 19.95 & 19.81
           & 17.29 & 18.04 & 18.07
           & 18.60 & 21.13 & 20.88 \\
\textbf{FKGL $\downarrow$}
  & Round 1    & 13.37 & 16.55 & 14.12
           & 13.92 & 17.51 & 15.51
           & 12.64 & 15.66 & 13.08
           & 12.57 & 16.88 & 14.48 \\
  & Round 2
           & \textbf{10.94} & \textbf{15.21} & \textbf{12.46}
           & \textbf{11.58} & \textbf{15.98} & \textbf{13.72}
           & \textbf{10.39} & \textbf{14.60} & \textbf{11.16}
           & \textbf{10.27} & \textbf{14.62} & \textbf{12.38} \\
\midrule

  & Initial Summary  & 14.25 & 13.69 & 15.04
           & 13.33 & 13.43 & 14.24
           & 13.11 & 12.77 & 14.14
           & 13.76 & 13.36 & 15.10 \\
\textbf{DCRS $\downarrow$}
  & Round 1    & 13.21 & 12.67 & 13.61
           & 12.49 & 12.51 & 13.31
           & 12.06 & 11.88 & 12.68
           & 12.31 & 12.14 & 13.32 \\
  & Round 2
           & \textbf{12.56} & \textbf{12.06} & \textbf{13.15}
           & \textbf{11.96} & \textbf{11.96} & \textbf{12.90}
           & \textbf{11.50} & \textbf{11.40} & \textbf{12.24}
           & \textbf{11.70} & \textbf{11.47} & \textbf{12.74} \\
\midrule

  & Initial Summary  & 18.72 & 17.05 & 20.86
           & 17.77 & 16.60 & 19.38
           & 16.56 & 15.13 & 17.38
           & 18.32 & 16.45 & 21.52 \\
\textbf{CLI $\downarrow$}
  & Round 1    & 15.86 & 14.54 & 16.54
           & 15.71 & 15.09 & 17.06
           & 14.07 & 13.59 & 14.22
           & 14.69 & 13.76 & 16.29 \\
  & Round 2
           & \textbf{13.96} & \textbf{13.20} & \textbf{15.14}
           & \textbf{14.20} & \textbf{14.23} & \textbf{16.01}
           & \textbf{12.56} & \textbf{12.62} & \textbf{13.09}
           & \textbf{12.84} & \textbf{12.42} & \textbf{14.71} \\
\midrule

  & Initial Summary  & 65.38 & 64.80 & 62.49
           & 60.28 & 55.37 & 56.04
           & 60.91 & 58.38 & 56.29
           & 59.62 & 55.28 & 55.86 \\
\textbf{LENS}
  & Round 1    & 71.35 & \textbf{67.21} & 71.62
           & 66.32 & 59.20 & 63.37
           & 68.20 & 62.58 & 66.68
           & 70.51 & 61.08 & 69.34 \\
  & Round 2
           & \textbf{74.67} & 66.68 & \textbf{72.60}
           & \textbf{70.00} & \textbf{60.60} & \textbf{64.61}
           & \textbf{71.63} & \textbf{62.68} & \textbf{68.30}
           & \textbf{74.44} & \textbf{62.62} & \textbf{71.56} \\
\midrule

  & Initial Summary  & 55.48 & 49.67 & 44.48
           & 42.36 & 43.63 & 38.59
           & 34.41 & 33.48 & 30.69
           & 53.19 & 54.73 & 42.87 \\
\textbf{SummaC}
  & Round 1    & 67.12 & 49.17 & 48.88
           & 53.14 & 45.70 & 45.38
           & 39.41 & 34.51 & 36.02
           & 69.53 & 57.95 & 49.87 \\
  & Round 2
           & \textbf{72.93} & \textbf{50.86} & \textbf{53.81}
           & \textbf{61.27} & \textbf{47.76} & \textbf{49.00}
           & \textbf{41.56} & \textbf{34.55} & \textbf{38.50}
           & \textbf{73.49} & \textbf{58.94} & \textbf{54.58} \\
\bottomrule

\end{tabular}

\caption{Performance of three models (GPT-4o, Llama-3.1-8B-Instruct, Qwen3-8B) on four benchmarks across initial summary and two refinement rounds; best scores in bold. ↓ indicates that lower is better in readability metrics.}
\label{tab:full-key-results}
\end{table*}

\subsection{Models}
Without additional fine-tuning, we evaluate NRLB using three large language models with distinct characteristics. Llama-3.1-8B-Instruct, an open-source model licensed under Apache 2.0, serves as a reproducible open-source baseline \cite{dubey2024llama}. GPT-4o represents OpenAI’s recent commercial model with multimodal capabilities \cite{hurst2024gpt}. Qwen3-8B in reasoning mode is included for its strong performance on complex reasoning tasks \cite{yang2025qwen3}. Full configuration details are provided in Appendix~\ref{sec:C2}.

\subsection{Metrics}
We evaluate NRLB across three core dimensions as defined in the Plain Language Summarization Shared Task: Relevance, Readability, and Factuality \cite{goldsack2024overview}. Relevance measures how well the summary preserves key content, Readability reflects how easily the summary can be understood by readers, and Factuality evaluates the consistency of the summary with the source. From the available metrics, we select the seven most aligned with our task: ROUGE-1, BERTScore, FKGL, DCRS, CLI, LENS, and SummaC.

Relevance is measured using ROUGE-1, which captures unigram overlap between summaries and references \cite{liu2008correlation}, and BERTScore, which quantifies semantic similarity using contextual embeddings \cite{zhang2019bertscore}. Readability is assessed through traditional formula-based metrics (FKGL, DCRS, and CLI) that reflect sentence and word complexity \cite{tanprasert2021flesch, dale1948formula,coleman1975computer}, along with LENS, a learned metric that better captures qualitative improvements from lexical and structural simplification \cite{maddela2022lens}. Factuality is evaluated using SummaC, which applies natural language inference to assess consistency between the source document and summary \cite{laban2022summac}. Detailed descriptions are provided in Appendix~\ref{sec:C3}.

\begin{table*}[t]
\centering
\scriptsize  
\setlength{\tabcolsep}{0.9pt}
\renewcommand{\arraystretch}{0.85}  
\begin{tabular}{lcccc|cccc|cccc|cccc}
\toprule
\multirow{2}{*}{\textbf{Method}}
  & \multicolumn{4}{c|}{\textbf{PLOS}}
  & \multicolumn{4}{c|}{\textbf{GovReport}}
  & \multicolumn{4}{c|}{\textbf{BillSum}}
  & \multicolumn{4}{c}{\textbf{BigPatent}} \\
\cmidrule(lr){2-5}\cmidrule(lr){6-9}\cmidrule(lr){10-13}\cmidrule(lr){14-17}
  & \textbf{FKGL $\downarrow$} & \textbf{DCRS $\downarrow$} & \textbf{CLI $\downarrow$} & \textbf{LENS}
  & \textbf{FKGL $\downarrow$} & \textbf{DCRS $\downarrow$} & \textbf{CLI $\downarrow$} & \textbf{LENS}
  & \textbf{FKGL $\downarrow$} & \textbf{DCRS $\downarrow$} & \textbf{CLI $\downarrow$} & \textbf{LENS}
  & \textbf{FKGL $\downarrow$} & \textbf{DCRS $\downarrow$} & \textbf{CLI $\downarrow$} & \textbf{LENS} \\
\midrule
All             
  & 15.40 & 12.06 & 13.19 & \textbf{67.64} 
  & \textbf{16.09} & \textbf{12.01} & \textbf{14.29} & 59.08 
  & 15.00 & 11.68 & 13.09 & 62.63 
  & \textbf{14.68} & \textbf{11.51} & \textbf{12.93} & \textbf{64.40} \\
w/o ele         
  & 15.73 & 12.19 & 13.45 & 66.18 
  & 16.56 & 12.05 & 14.68 & 59.94 
  & \textbf{14.48} & \textbf{11.35} & \textbf{12.52} & 62.26 
  & 15.23 & 11.67 & 12.79 & 61.90 \\
w/o non         
  & \textbf{15.20} & \textbf{12.01} & 13.10 & 66.36 
  & 16.19 & \textbf{12.01} & 14.41 & \textbf{60.12} 
  & 14.52 & 11.47 & 12.60 & 62.10

  & 16.83 & 11.91 & 13.31 & 59.82 \\
w/o att         
  & 15.40 & 12.06 & 13.21 & 67.31 
  & 16.34 & 12.06 & 14.51 & 59.27 
  & 14.49 & 11.42 & 12.58 & \textbf{63.01} 
  & 15.57 & 11.74 & 13.00 & 62.34 \\
w/o ele,non     
  & 16.01 & 12.28 & 13.68 & 64.75 
  & 17.03 & 12.05 & 14.88 & 58.26 
  & 14.61 & 11.39 & 12.58 & 62.77 
  & 16.64 & 11.90 & 13.31 & 58.71 \\
w/o ele,att     
  & 15.74 & 12.24 & 13.47 & 64.84 
  & 16.83 & 12.13 & 14.81 & 58.20 
  & 14.73 & 11.43 & 12.56 & 62.09 
  & 17.79 & 12.08 & 13.73 & 57.91 \\
w/o non,att     
  & 15.27 & 12.03 & \textbf{13.09} & 66.84 
  & 16.23 & 12.07 & 14.42 & 58.74 
  & 14.72 & 11.49 & 12.72 & 62.09 
  & 16.15 & 11.88 & 13.26 & 61.71 \\
\bottomrule
\end{tabular}
\caption{Ablation study on the effect of Reader Agent combinations across four datasets (200 samples each). “ele” = Elementary School Student Reader, “non” = Non-Native Reader, “att” = Attention-Deficit Reader.}
\label{tab:ablation_all}
\end{table*}

\section{Results}
\subsection{Improving Readability via Iterative Revision}
Table~\ref{tab:full-key-results} summarizes the effects of two rounds of revision (Rounds 1 and 2) using NRLB across four datasets and three models. Iterative refinement consistently improved readability across all datasets. In contrast, ROUGE-1 and BERTScore declined slightly due to reduced lexical overlap, reflecting the well-known trade-off in plain language summarization where clarity often reduces surface similarity \cite{dreyer2021evaluating, goldsack2024overview}. Meanwhile, factuality improved across all settings, with SummaC scores increasing from 55.48 to 72.93 on PLOS and from 53.19 to 73.49 on BigPatent, confirming that iterative revision enhanced both logical consistency and source alignment.

Across models, GPT-4o showed the best overall balance, maintaining strong initial relevance and achieving the highest readability and factuality by Round 2. Llama-3.1-8B-Instruct retained high ROUGE but made modest gains elsewhere. Qwen3-8B, though initially less consistent, achieved the largest improvements after revision, likely due to its reasoning-oriented architecture. While Round 3 results (Appendix~\ref{sec:D1}) brought minor further improvements in readability and factuality, relevance dropped by up to 10 points compared to the initial summary, supporting Round 2 as the default stopping point. This behavior reflects a relevance-readability trade-off: aggressive simplification in later rounds tends to disrupt semantic flow and reduce lexical alignment with the source. Based on this empirical observation, we adopt Round 2 as the default setting, balancing accessibility with content preservation. Efficiency analysis, including latency, API calls, and average cost, is provided in Appendix~\ref{sec:D2}.

\begin{table}[t]
\centering
\scriptsize
\renewcommand{\arraystretch}{1.2}
\setlength{\tabcolsep}{3pt}
\begin{tabular}{llccc}
\toprule
\textbf{Dataset} & \textbf{Setting} & \textbf{Initial} & \textbf{Round 1} & \textbf{Round 2} \\
\midrule
\multirow{2}{*}{PLOS} 
  & w/ Editor     & 49.6662 & 49.1664 (\textit{$-$0.50})         & \textbf{52.3063} (\textbf{\textit{+3.14}}) \\
  & w/o Editor    & 49.9958 & 48.8337 (\textit{$-$1.16})         & 49.8839 (\textbf{\textit{+1.05}}) \\
\midrule
\multirow{2}{*}{GovReport} 
  & w/ Editor    & 43.6340 & 45.7042 (\textbf{\textit{+2.07}})  & \textbf{49.9039} (\textbf{\textit{+4.20}}) \\
  & w/o Editor   & 43.6733 & 43.0163 (\textit{$-$0.66})         & 43.6692 (\textbf{\textit{+0.65}}) \\
\midrule
\multirow{2}{*}{BillSum} 
  & w/ Editor     & 33.4814 & 34.5100 (\textbf{\textit{+1.03}})  & \textbf{35.0046} (\textbf{\textit{+0.49}}) \\
  & w/o Editor    & 33.2260 & 33.4913 (\textbf{\textit{+0.27}})  & 33.2915 (\textit{$-$0.20}) \\
\midrule
\multirow{2}{*}{BigPatent} 
  & w/ Editor    & 54.7311 & 57.9544 (\textbf{\textit{+3.22}})  & \textbf{58.9384} (\textbf{\textit{+0.98}}) \\
  & w/o Editor  & 55.2051 & 55.0744 (\textit{$-$0.13})         & 55.8485 (\textbf{\textit{+0.77}}) \\
\bottomrule
\end{tabular}
\caption{SummaC scores across datasets and revision rounds (Initial denotes the initial summary) with and without the Editor (Editor Agent). Positive changes are shown in bold.}
\label{tab:editor_ablation}
\end{table}

\subsection{Ablation Study}
\label{sec:ablation}
\paragraph{Effect of Reader Agent Configurations}
We evaluated the effect of using one, two, and three Reader Agents. As shown in Table~\ref{tab:ablation_all}, all configurations were evaluated under the same two-round revision pipeline using four readability metrics. Using two Reader Agents generally improved readability across datasets and metrics, resulting in improved clarity and accessibility. For instance, on GovReport, FKGL improved from 17.03 to 16.09, and LENS rose from 58.20 to 60.12.

With three Reader Agents, FKGL improved further and LENS reached its highest scores on PLOS and BigPatent, indicating the strongest overall improvements. In contrast, BillSum achieved near-optimal readability with two Reader Agents, and the third added only marginal gains. These findings suggest that using at least two Reader Agents is generally beneficial for readability, with the three-agent setup performing best overall. We therefore adopt the three-agent configuration as the default. Additional results with four and five agents are reported in Appendix~\ref{sec:B1}, and further analysis on feedback overlap and the complementary roles of Reader Agents is provided in Appendix~\ref{sec:B2}.

\paragraph{Effect of the Editor Agent} As shown in Table~\ref{tab:editor_ablation}, the inclusion of the Editor Agent consistently improves SummaC, a factuality metric, at each revision round, yielding final values well above those without the Editor Agent. In contrast, without the Editor Agent, the scores show inconsistent changes in Round 1, with both increases and decreases observed, and this instability persists in Round 2. These results indicate that the Editor Agent is essential for merging feedback from Reader Agents and the Domain Expert coherently, producing steady gains in factuality and improving the overall reliability of the final summary. In particular, the Editor Agent plays a central role in transforming multi-agent feedback into actionable revisions, thereby improving overall performance.

\begin{figure*}[t]
    \centering
    \includegraphics[scale=0.32]{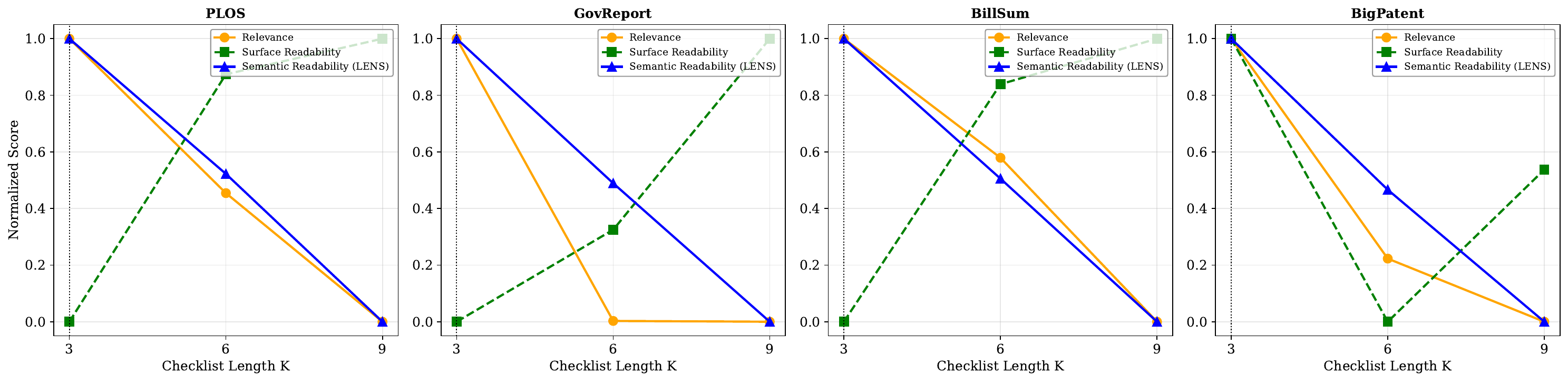}
    \caption{Impact of checklist length on summary quality; K = 3 achieves the best trade-off between relevance and readability.}
    \label{fig:checklist}
\end{figure*}

\begin{table*}[t]
\centering
\small
\renewcommand{\arraystretch}{0.90}
\setlength{\tabcolsep}{0.90pt}
\resizebox{1.0\textwidth}{!}{%

\begin{tabular*}{\textwidth}{@{\extracolsep{\fill}} l l c c c c c c c}
\toprule
& & \multicolumn{2}{c}{\textbf{Relevance}} 
  & \multicolumn{4}{c}{\textbf{Readability}} 
  & \textbf{Factuality} \\
\cmidrule(lr){3-4} \cmidrule(lr){5-8} \cmidrule(lr){9-9}
\textbf{Dataset} & \textbf{Method} 
  & \textbf{ROUGE-1} & \textbf{BERTScore} 
  & \textbf{FKGL $\downarrow$} & \textbf{DCRS $\downarrow$} & \textbf{CLI $\downarrow$} & \textbf{LENS} 
  & \textbf{SummaC} \\
\midrule
\multirow{3}{*}{PLOS (500 samples)}
  & AgentSimp-Synchronous   & 44.59 & 86.23 & 22.20 & 13.85 & 16.45 & 60.08 & 45.44 \\
  & AgentSimp-Pipeline      & \textbf{44.92} & \textbf{86.31} & 22.33 & 13.84 & 16.37 & 60.79 & 46.07 \\
  & \textbf{\emph{NRLB (Ours)}} 
                           & 41.64 & 85.43 & \textbf{15.21} & \textbf{12.06} & \textbf{13.20} & \textbf{66.68} & \textbf{50.86} \\
\midrule
\multirow{3}{*}{GovReport (500 samples)}
  & AgentSimp-Synchronous   & 43.62 & 85.55 & 27.41 & 14.44 & 17.14 & 50.24 & 45.14 \\
  & AgentSimp-Pipeline      & 43.39 & 85.56 & 26.31 & 14.28 & 17.08 & 50.62 & 45.63 \\
  & \textbf{\emph{NRLB (Ours)}} 
                           & \textbf{45.33} & \textbf{85.84} & \textbf{17.51} & \textbf{12.51} & \textbf{15.09} & \textbf{59.20} & \textbf{45.70} \\
\midrule
\multirow{3}{*}{BillSum (500 samples)}
  & AgentSimp-Synchronous   & 37.82 & 85.31 & 26.36 & 12.99 & 14.37 & 56.47 & 32.86 \\
  & AgentSimp-Pipeline      & 34.11 & 84.83 & 50.08 & 15.95 & 14.69 & 46.77 & 30.67 \\
  & \textbf{\emph{NRLB (Ours)}} 
                           & \textbf{39.70} & \textbf{85.59} & \textbf{14.60} & \textbf{11.40} & \textbf{12.62} & \textbf{62.68} & \textbf{34.55} \\
\midrule
\multirow{3}{*}{BigPatent (500 samples)}
  & AgentSimp-Synchronous   & 37.38 & 85.44 & 23.30 & 13.10 & 14.28 & 53.47 & 48.27 \\
  & AgentSimp-Pipeline      & \textbf{37.45} & \textbf{85.46} & 23.96 & 13.14 & 14.14 & 53.08 & 48.24 \\
  & \textbf{\emph{NRLB (Ours)}} 
                           & 36.87 & 85.11 & \textbf{14.62} & \textbf{11.47} & \textbf{12.42} & \textbf{62.62} & \textbf{58.94} \\
\bottomrule
\end{tabular*}
} 
\caption{Automatic evaluation of NRLB and AgentSimp variants on four benchmarks (500 samples each). Best scores are shown in bold.}
\label{tab:simplification_comparison}
\end{table*}

\subsection{Effect of Checklist Size}

We study the impact of the number of feedback items (K) on summary quality, comparing K = 3, 6, and 9. Results are shown in Figure~\ref{fig:checklist}. To analyze trade-offs, we report three normalized metrics for each dataset: Relevance (average of ROUGE-1 and BERTScore), Surface Readability (inverse average of FKGL, DCRS, and CLI), and Semantic Readability (LENS).

As K increases, surface readability improves, reflecting simpler syntax and shorter sentences. However, both Relevance and LENS decline, indicating loss of key content and degraded semantic coherence. These results suggest that larger checklists tend to over-simplify summaries and reduce informativeness. We therefore adopt K = 3 as the default setting, which provides the best overall balance. For completeness, results for K = 1 and K = 2 are reported in Table~\ref{tab:checklist}, both yielding lower readability and factuality than K = 3.

\subsection{Comparison with Baselines}
We compare NRLB’s adaptive feedback loop with two communication strategies from AgentSimp \cite{fang2025collaborative}: synchronous and pipeline-style settings. For a fair comparison, all systems were evaluated under a single-pass configuration (one revision round). NRLB achieved the strongest overall readability performance across all four datasets, improving FKGL, DCRS, CLI, and LENS scores. NRLB also achieved the highest factuality scores, reflecting stronger coherence and alignment with the source. In terms of relevance, NRLB outperformed AgentSimp on GovReport and BillSum, but showed slight decreases on PLOS and BigPatent, suggesting a trade-off between readability and content preservation. Overall, NRLB achieved the most balanced improvements even under a single-pass setup. 

We additionally examined a direct prompting baseline using the same underlying LLM, but without agent collaboration or iterative refinement. In this setting, the model performs single-pass simplification based on a simple instruction prompt. This approach aggressively removes information during simplification, improving readability metrics (FKGL, DCRS, and CLI) but consistently reducing relevance (ROUGE-1 and BERTScore). These results suggest that excessive simplification can enhance surface-level readability at the cost of losing essential content. As a consequence, the generated summaries often exhibit limited informational coverage.
As a result, the direct baseline exhibits behavior that differs from our goal of improving readability while preserving key information. We therefore exclude it from the main comparison and report its implementation details and additional results in Appendix~\ref{sec:C4}.

\begin{table}[t]
\centering
\scriptsize
\setlength{\tabcolsep}{1.5pt}  
\renewcommand{\arraystretch}{0.9}  
\begin{tabular}{llccc|ccc|c}
\toprule
\multirow{2}{*}{\textbf{Group}} & \multirow{2}{*}{\textbf{Dataset}} &
\multicolumn{3}{c|}{\textbf{NRLB}} &
\multicolumn{3}{c|}{\textbf{AgentSimp}} &
\multirow{2}{*}{\textbf{Pref. (\%)}} \\
\cmidrule{3-8}
& & Coh. & Simpl. & Faith. & Coh. & Simpl. & Faith. & \\
\midrule
\multirow{6}{*}{Non-Native} 
& PLOS    & \textbf{4.10} & \textbf{3.85} & \textbf{3.85} & 3.77 & 2.61 & 3.78 & 75.0 \\
& GovR.   & \textbf{4.27} & \textbf{3.73} & \textbf{4.12} & 3.72 & 3.35 & 3.62 & 65.0 \\
& BillSum & \textbf{4.30} & \textbf{3.88} & \textbf{3.97} & 3.37 & 2.97 & 3.40 & 71.7 \\
& Patent  & \textbf{4.27} & \textbf{4.18} & \textbf{4.12} & 3.40 & 2.35 & 3.32 & 76.7 \\
\cmidrule(lr){2-9}
& Average & 4.23 & 3.91 & 4.00 & 3.56 & 2.82 & 3.53 & 72.1 \\
\midrule
\multirow{6}{*}{\parbox{1.5cm}{Elementary\\School\\Student}}
& PLOS    & 3.17 & 2.83 & \textbf{3.33} & \textbf{3.33} & 2.83 & 2.83 & 66.7 \\
& GovR.   & \textbf{3.89} & \textbf{3.56} & \textbf{3.33} & 3.66 & 3.11 & 3.00 & 55.7 \\
& BillSum & \textbf{3.83} & 2.83 & \textbf{3.50} & 3.17 & \textbf{3.17} & 3.17 & 67.0 \\
& Patent  & \textbf{3.78} & \textbf{3.33} & \textbf{3.89} & 3.44 & 3.00 & 3.33 & 56.0 \\
\cmidrule(lr){2-9}
& Average & 3.70 & 3.20 & 3.53 & 3.43 & 3.03 & 3.10 & 61.5 \\
\bottomrule
\end{tabular}
\caption{Human evaluation results (5-point Likert scale) for NRLB vs. AgentSimp across two groups (Non-Native and Elementary School Student). “GovR.” denotes GovReport; bold indicates the higher score.}
\label{tab:human_eval_combined}
\end{table}

\begin{figure}[t]
    \centering
    \includegraphics[scale=0.65]{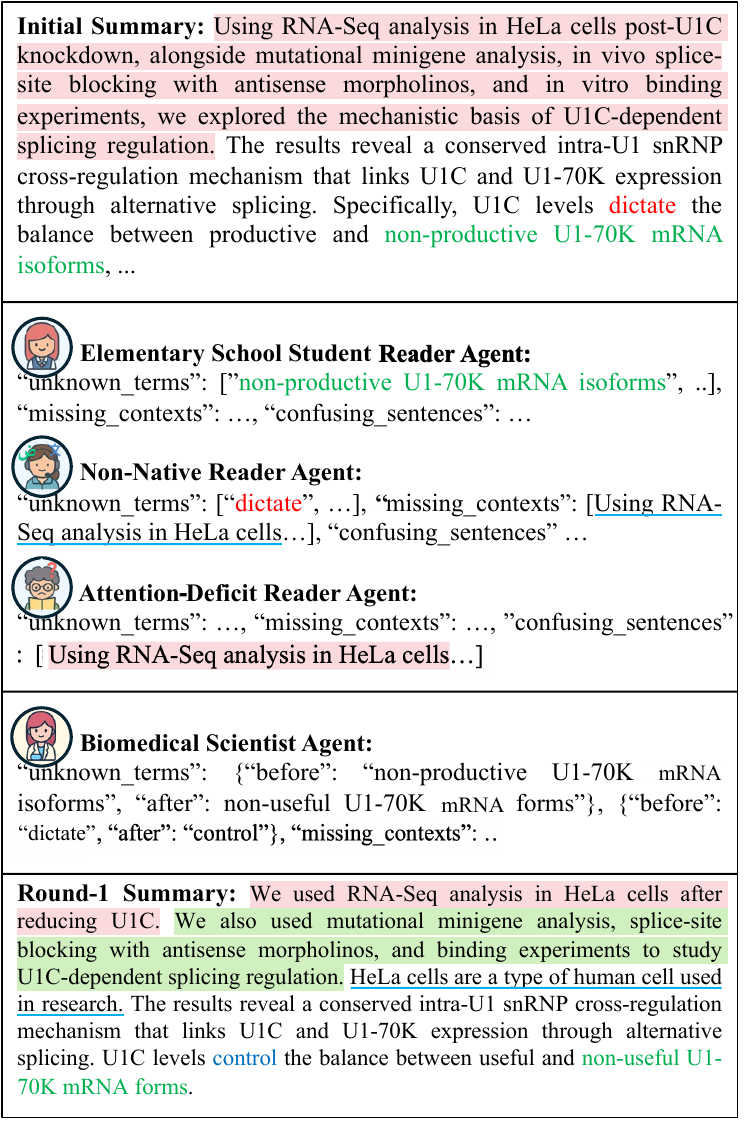}
    \caption{Example case study shows how reader feedback drives lexical, contextual, and structural simplifications in NRLB.}
    \label{fig:qualitative}
\end{figure}

\subsection{Human Evaluation and Qualitative Analysis}
We conduct a human evaluation with two participant groups to compare NRLB with AgentSimp (see Table~\ref{tab:human_eval_combined}). First, three undergraduate non-native English speakers with intermediate English proficiency evaluated 80 documents. Each annotator rated the summaries on coherence, simplicity, and faithfulness using a five-point scale and selected the version they found easier to understand. NRLB achieved consistently higher scores across all metrics, with average improvements of 0.49-1.09 points. In preference judgments, 65-77\% of cases (72.1\% on average) favored NRLB, confirming its benefits for readability and user satisfaction. 

We also conducted an evaluation with three elementary school students with limited English proficiency on 10 documents. NRLB again outperformed AgentSimp across most metrics, showing particularly clear improvements in coherence and faithfulness for GovReport, BillSum, and BigPatent. However, AgentSimp slightly outperformed NRLB on coherence in PLOS and on simplicity in BillSum. Overall, NRLB was preferred in 61.5\% of cases, demonstrating its effectiveness in reducing reading difficulty and preserving essential content across diverse user groups. Further details of the human evaluation, including inter-annotator agreement analysis, are provided in Appendix~\ref{sec:E}. Evaluation for the Attention-Deficit Reader group was not conducted due to ethical and clinical constraints, and the elementary-student evaluation was limited in scale to comply with IRB requirements.

Figure~\ref{fig:qualitative} presents a representative case study that exemplifies these improvements. Additional case studies and failure case analyses are provided in Appendix~\ref{sec:F} and Appendix~\ref{sec:G}.

\section{Conclusion}
This paper presents NRLB (No Reader Left Behind), a multi-agent framework for generating plain language summaries by simulating diverse reader perspectives. The system combines genre-specific content planning with iterative feedback and revision, guided by Reader, Checklist, Domain Expert, and Editor Agents. Across four datasets, NRLB consistently improved readability while improving factual consistency and preserving key content through collaborative refinement. Human evaluations confirmed that NRLB produces summaries that are easier to understand, more coherent, and better suited for diverse reader groups. These results demonstrate the practical applicability of NRLB as a viable approach to implementing plain language policies such as the U.S. Plain Writing Act. Future work includes extending NRLB to broader domains and incorporating more diverse reader profiles.

\section*{Limitations}

While the current implementation of the NRLB pipeline demonstrates the feasibility of multi-agent readability-guided summarization, several limitations remain.

First, readability feedback relies on large language models that simulate elementary school students, non-native readers, and readers with attention deficits. Although the prompts are designed to elicit diverse comprehension challenges, the responses are not grounded in real-world reader data and may include hallucinated or redundant feedback, particularly in edge cases.

Second, human evaluation is limited in scale and coverage due to ethical and practical constraints. The evaluation with elementary school students was restricted to a subset of documents to comply with IRB requirements, while evaluation for the attention-deficit reader group was not conducted due to the need for clinical validation and handling sensitive personal health information, limiting generalizability and requiring large-scale evaluation.

Third, iterative refinement may introduce a trade-off between readability and content preservation. As the number of revision rounds or feedback items increases, the system may over-simplify the text, leading to loss of important details or degradation of semantic coherence.

\section*{Acknowledgments}
This work was supported by the Commercialization Promotion Agency for R\&D Outcomes(COMPA) grant funded by the Korea government(Ministry of Science and ICT)(2710086166), Institute for Information \& communications Technology Promotion(IITP) grant funded by the Korea government(MSIT) (RS-2024-00398115, Research on the reliability and coherence of outcomes produced by Generative AI), Institute for Information \& communications Technology Planning \& Evaluation(IITP) grant funded by the Korea government(MSIT) (No. RS-2022-II220369, (Part 4) Development of AI Technology to support Expert Decision-making that can Explain the Reasons/Grounds for Judgment Results based on Expert Knowledge), and Institute of Information \& communications Technology Planning \& Evaluation (IITP) under the artificial intelligence star fellowship support program to nurture the best talents (IITP-2026-RS-2025-02304828) grant funded by the Korea government(MSIT).


\bibliography{custom}

@article{qin2024mass,
  title={The mass public’s science literacy and co-production during the COVID-19 pandemic: empirical evidence from 140 cities in China},
  author={Qin, Haibo and Xie, Zhongxuan and Shang, Huping and Sun, Yong and Yang, Xiaohui and Li, Mengming},
  journal={Humanities and Social Sciences Communications},
  volume={11},
  number={1},
  pages={1--13},
  year={2024},
  publisher={Palgrave}
}

@inproceedings{mo2024expertease,
  title={ExpertEase: A Multi-Agent Framework for Grade-Specific Document Simplification with Large Language Models},
  author={Mo, Kaijie and Hu, Renfen},
  booktitle={Findings of the Association for Computational Linguistics: EMNLP 2024},
  pages={9080--9099},
  year={2024}
}

@inproceedings{fang2025collaborative,
  title={Collaborative document simplification using multi-agent systems},
  author={Fang, Dengzhao and Qiang, Jipeng and Ouyang, Xiaoye and Zhu, Yi and Yuan, Yunhao and Li, Yun},
  booktitle={Proceedings of the 31st International Conference on Computational Linguistics},
  pages={897--912},
  year={2025}
}

@article{du2024multi,
  title={Multi-agent software development through cross-team collaboration},
  author={Du, Zhuoyun and Qian, Chen and Liu, Wei and Xie, Zihao and Wang, Yifei and Dang, Yufan and Chen, Weize and Yang, Cheng},
  journal={arXiv preprint arXiv:2406.08979},
  year={2024}
}

@article{laban2022summac,
  title={SummaC: Re-visiting NLI-based models for inconsistency detection in summarization},
  author={Laban, Philippe and Schnabel, Tobias and Bennett, Paul N and Hearst, Marti A},
  journal={Transactions of the Association for Computational Linguistics},
  volume={10},
  pages={163--177},
  year={2022},
  publisher={MIT Press One Rogers Street, Cambridge, MA 02142-1209, USA journals-info~…}
}

@inproceedings{wilkens2020simplifying,
  title={Simplifying coreference chains for dyslexic children},
  author={Wilkens, Rodrigo and Todirascu, Amalia},
  booktitle={Proceedings of the Twelfth Language Resources and Evaluation Conference},
  pages={1142--1151},
  year={2020}
}

@article{xu2025evaluating,
  title={Evaluating small language models for news summarization: Implications and factors influencing performance},
  author={Xu, Borui and Chen, Yao and Wen, Zeyi and Liu, Weiguo and He, Bingsheng},
  journal={arXiv preprint arXiv:2502.00641},
  year={2025}
}

@article{goldsack2022making,
  title={Making science simple: Corpora for the lay summarisation of scientific literature},
  author={Goldsack, Tomas and Zhang, Zhihao and Lin, Chenghua and Scarton, Carolina},
  journal={arXiv preprint arXiv:2210.09932},
  year={2022}
}

@article{huang2021efficient,
  title={Efficient attentions for long document summarization},
  author={Huang, Luyang and Cao, Shuyang and Parulian, Nikolaus and Ji, Heng and Wang, Lu},
  journal={arXiv preprint arXiv:2104.02112},
  year={2021}
}

@article{kornilova2019billsum,
  title={BillSum: A corpus for automatic summarization of US legislation},
  author={Kornilova, Anastassia and Eidelman, Vlad},
  journal={arXiv preprint arXiv:1910.00523},
  year={2019}
}

@article{sharma2019bigpatent,
  title={BIGPATENT: A large-scale dataset for abstractive and coherent summarization},
  author={Sharma, Eva and Li, Chen and Wang, Lu},
  journal={arXiv preprint arXiv:1906.03741},
  year={2019}
}

@article{hurst2024gpt,
  title={Gpt-4o system card},
  author={Hurst, Aaron and Lerer, Adam and Goucher, Adam P and Perelman, Adam and Ramesh, Aditya and Clark, Aidan and Ostrow, AJ and Welihinda, Akila and Hayes, Alan and Radford, Alec and others},
  journal={arXiv preprint arXiv:2410.21276},
  year={2024}
}

@article{yang2025qwen3,
  title={Qwen3 technical report},
  author={Yang, An and Li, Anfeng and Yang, Baosong and Zhang, Beichen and Hui, Binyuan and Zheng, Bo and Yu, Bowen and Gao, Chang and Huang, Chengen and Lv, Chenxu and others},
  journal={arXiv preprint arXiv:2505.09388},
  year={2025}
}

@article{dubey2024llama,
  title={The llama 3 herd of models},
  author={Dubey, Abhimanyu and Jauhri, Abhinav and Pandey, Abhinav and Kadian, Abhishek and Al-Dahle, Ahmad and Letman, Aiesha and Mathur, Akhil and Schelten, Alan and Yang, Amy and Fan, Angela and others},
  journal={arXiv e-prints},
  pages={arXiv--2407},
  year={2024}
}

@article{goldsack2024overview,
  title={Overview of the biolaysumm 2024 shared task on the lay summarization of biomedical research articles},
  author={Goldsack, Tomas and Scarton, Carolina and Shardlow, Matthew and Lin, Chenghua},
  journal={arXiv preprint arXiv:2408.08566},
  year={2024}
}

@inproceedings{liu2008correlation,
  title={Correlation between rouge and human evaluation of extractive meeting summaries},
  author={Liu, Feifan and Liu, Yang},
  booktitle={Proceedings of ACL-08: HLT, short papers},
  pages={201--204},
  year={2008}
}

@article{zhang2019bertscore,
  title={Bertscore: Evaluating text generation with bert},
  author={Zhang, Tianyi and Kishore, Varsha and Wu, Felix and Weinberger, Kilian Q and Artzi, Yoav},
  journal={arXiv preprint arXiv:1904.09675},
  year={2019}
}

@article{fabbri2021summeval,
  title={Summeval: Re-evaluating summarization evaluation},
  author={Fabbri, Alexander R and Kry{\'s}ci{\'n}ski, Wojciech and McCann, Bryan and Xiong, Caiming and Socher, Richard and Radev, Dragomir},
  journal={Transactions of the Association for Computational Linguistics},
  volume={9},
  pages={391--409},
  year={2021},
  publisher={MIT Press One Rogers Street, Cambridge, MA 02142-1209, USA journals-info~…}
}

@article{maddela2022lens,
  title={LENS: A learnable evaluation metric for text simplification},
  author={Maddela, Mounica and Dou, Yao and Heineman, David and Xu, Wei},
  journal={arXiv preprint arXiv:2212.09739},
  year={2022}
}

@inproceedings{tanprasert2021flesch,
  title={Flesch-kincaid is not a text simplification evaluation metric},
  author={Tanprasert, Teerapaun and Kauchak, David},
  booktitle={Proceedings of the 1st Workshop on Natural Language Generation, Evaluation, and Metrics (GEM 2021)},
  pages={1--14},
  year={2021}
}

@article{coleman1975computer,
  title={A computer readability formula designed for machine scoring.},
  author={Coleman, Meri and Liau, Ta Lin},
  journal={Journal of Applied Psychology},
  volume={60},
  number={2},
  pages={283},
  year={1975},
  publisher={American Psychological Association}
}

@article{mahapatra2024extensive,
  title={An Extensive Evaluation of Factual Consistency in Large Language Models for Data-to-Text Generation},
  author={Mahapatra, Joy and Garain, Utpal},
  journal={arXiv preprint arXiv:2411.19203},
  year={2024}
}

@article{laufer2010lexical,
  title={Lexical threshold revisited: Lexical text coverage, learners’ vocabulary size and reading comprehension},
  author={Laufer, Batia and Ravenhorst‑Kalovski, Geke C.},
  journal={Reading in a Foreign Language},
  volume={22},
  number={1},
  pages={15--30},
  year={2010},
  issn={1539-0578}
}

@article{ha2022vocabulary,
  title={Vocabulary demands of informal spoken English revisited: what does it take to understand movies, TV programs, and soap operas?},
  author={Ha, Hung Tan},
  journal={Frontiers in Psychology},
  volume={13},
  pages={831684},
  year={2022},
  publisher={Frontiers Media SA}
}

@article{jacobson2011working,
  title={Working memory influences processing speed and reading fluency in ADHD},
  author={Jacobson, Lisa A and Ryan, Matthew and Martin, Rebecca B and Ewen, Joshua and Mostofsky, Stewart H and Denckla, Martha B and Mahone, E Mark},
  journal={Child neuropsychology},
  volume={17},
  number={3},
  pages={209--224},
  year={2011},
  publisher={Taylor \& Francis}
}

@techreport{senter1967automated,
  title={Automated Readability Index},
  author={Senter, R.\,J. and Smith, E.\,A.},
  institution={Aerospace Medical Division, Aerospace Medical Research Laboratories, Wright‑Patterson Air Force Base, Ohio},
  number={AMRL‑TR‑66‑22},
  year={1967},
  note={U.S. Air Force Systems Command, Project 1710 (Human Factors in the Design of Training Systems), Task 171007}
}

@article{gooding2022ethical,
  title={On the ethical considerations of text simplification},
  author={Gooding, Sian},
  journal={arXiv preprint arXiv:2204.09565},
  year={2022}
}

@article{xu2015problems,
  title={Problems in current text simplification research: New data can help},
  author={Xu, Wei and Callison-Burch, Chris and Napoles, Courtney},
  journal={Transactions of the Association for Computational Linguistics},
  volume={3},
  pages={283--297},
  year={2015},
  publisher={MIT Press One Rogers Street, Cambridge, MA 02142-1209, USA journals-info~…}
}

@article{luo2022readability,
  title={Readability controllable biomedical document summarization},
  author={Luo, Zheheng and Xie, Qianqian and Ananiadou, Sophia},
  journal={arXiv preprint arXiv:2210.04705},
  year={2022}
}

@article{urlana2023controllable,
  title={Controllable Text Summarization: Unraveling Challenges, Approaches, and Prospects--A Survey},
  author={Urlana, Ashok and Mishra, Pruthwik and Roy, Tathagato and Mishra, Rahul},
  journal={arXiv preprint arXiv:2311.09212},
  year={2023}
}

@article{chen2023agentverse,
  title={Agentverse: Facilitating multi-agent collaboration and exploring emergent behaviors in agents},
  author={Chen, Weize and Su, Yusheng and Zuo, Jingwei and Yang, Cheng and Yuan, Chenfei and Qian, Chen and Chan, Chi-Min and Qin, Yujia and Lu, Yaxi and Xie, Ruobing and others},
  journal={arXiv preprint arXiv:2308.10848},
  volume={2},
  number={4},
  pages={6},
  year={2023}
}

@article{dadu2021text,
  title={Text simplification for comprehension-based question-answering},
  author={Dadu, Tanvi and Pant, Kartikey and Nagar, Seema and Barbhuiya, Ferdous Ahmed and Dey, Kuntal},
  journal={arXiv preprint arXiv:2109.13984},
  year={2021}
}

@inproceedings{rello2013frequent,
  title={Frequent words improve readability and short words improve understandability for people with dyslexia},
  author={Rello, Luz and Baeza-Yates, Ricardo and Dempere-Marco, Laura and Saggion, Horacio},
  booktitle={IFIP Conference on Human-Computer Interaction},
  pages={203--219},
  year={2013},
  organization={Springer}
}

@article{smith2021role,
  title={The role of background knowledge in reading comprehension: A critical review},
  author={Smith, Reid and Snow, Pamela and Serry, Tanya and Hammond, Lorraine},
  journal={Reading Psychology},
  volume={42},
  number={3},
  pages={214--240},
  year={2021},
  publisher={Taylor \& Francis}
}

@article{tighe2016examining,
  title={Examining the relationships of component reading skills to reading comprehension in struggling adult readers: A meta-analysis},
  author={Tighe, Elizabeth L and Schatschneider, Christopher},
  journal={Journal of learning disabilities},
  volume={49},
  number={4},
  pages={395--409},
  year={2016},
  publisher={Sage Publications Sage CA: Los Angeles, CA}
}

@article{shero2021differential,
  title={The differential relations between ADHD and reading comprehension: A quantile regression and quantile genetic approach},
  author={Shero, Jeffrey A and Logan, Jessica AR and Petrill, Stephen A and Willcutt, Erik and Hart, Sara A},
  journal={Behavior genetics},
  volume={51},
  number={6},
  pages={631--653},
  year={2021},
  publisher={Springer}
}

@article{dreyer2021evaluating,
  title={Evaluating the tradeoff between abstractiveness and factuality in abstractive summarization},
  author={Dreyer, Markus and Liu, Mengwen and Nan, Feng and Atluri, Sandeep and Ravi, Sujith},
  journal={arXiv preprint arXiv:2108.02859},
  year={2021}
}

@article{dale1948formula,
  title={A formula for predicting readability: Instructions},
  author={Dale, Edgar and Chall, Jeanne S},
  journal={Educational research bulletin},
  pages={37--54},
  year={1948},
  publisher={JSTOR}
}

@article{argyle2023out,
  title={Out of one, many: Using language models to simulate human samples},
  author={Argyle, Lisa P and Busby, Ethan C and Fulda, Nancy and Gubler, Joshua R and Rytting, Christopher and Wingate, David},
  journal={Political Analysis},
  volume={31},
  number={3},
  pages={337--351},
  year={2023},
  publisher={Cambridge University Press}
}

@inproceedings{park2023generative,
  title={Generative agents: Interactive simulacra of human behavior},
  author={Park, Joon Sung and O'Brien, Joseph and Cai, Carrie Jun and Morris, Meredith Ringel and Liang, Percy and Bernstein, Michael S},
  booktitle={Proceedings of the 36th annual acm symposium on user interface software and technology},
  pages={1--22},
  year={2023}
}

@inproceedings{aher2023using,
  title={Using large language models to simulate multiple humans and replicate human subject studies},
  author={Aher, Gati V and Arriaga, Rosa I and Kalai, Adam Tauman},
  booktitle={International conference on machine learning},
  pages={337--371},
  year={2023},
  organization={PMLR}
}

\clearpage

\appendix
\section{Implementation Details}
\label{sec:A}
Our framework is built entirely on large language models (LLMs), enabling a training-free and easily deployable approach. The implementation centers on two key aspects. First is the design of the overall process, which generates an initial summary based on document type and agent roles, followed by iterative revision through multi-agent interactions. Second is the construction of optimized prompting templates tailored to each role (e.g., Domain Expert, Editor, Reader Agents). The workflow is outlined in Algorithm~\ref{alg:llm-summary-refine}.

\begin{algorithm}
\footnotesize
\caption{Multi-Agent Summarization with Multi-Round Readability Refinement}
\label{alg:llm-summary-refine}
\raggedright
\begin{algorithmic}[1]

\Require Source documents; prompts for Planner, Domain Expert, Readers, and Editor Agent
\Ensure Initial and $N$ refined summaries are written to file

\State Load source documents (JSON format) into $\mathsf{samples}$
\ForAll{sample $\in \mathsf{samples}$}
    \State Extract $\mathsf{source\_text}$

    \State $\mathsf{doc\_type}, \mathsf{expert} \leftarrow$ \Call{CallPlanner}{$\mathsf{source\_text}$}

    \State $\mathsf{initial\_summary} \leftarrow$ \Call{GenerateSummary}{}
    \Statex \hspace{\algorithmicindent} $(\mathsf{source\_text}, \mathsf{doc\_type}, \mathsf{expert})$

    \State Save $\mathsf{initial\_summary}$ to file
    \State Initialize Readers with role-specific prompts
    \State $\mathsf{curr} \leftarrow \mathsf{initial\_summary}$

    \For{$r = 1$ to $N$}
        \State $\mathsf{fb\_ele} \leftarrow$ \Call{CallReader}{ele, $\mathsf{curr}$}
        \State $\mathsf{fb\_non} \leftarrow$ \Call{CallReader}{non, $\mathsf{curr}$}
        \State $\mathsf{fb\_att} \leftarrow$ \Call{CallReader}{att, $\mathsf{curr}$}

        \State $\mathsf{edits} \leftarrow$ \Call{ChecklistAggregate}{}
        \Statex \hspace{\algorithmicindent} $(\mathsf{fb\_ele}, \mathsf{fb\_non}, \mathsf{fb\_att})$

        \State $\mathsf{sugg} \leftarrow$ \Call{CallDomainExpert}{$\mathsf{curr}, \mathsf{edits}$}
        \State $\mathsf{curr} \leftarrow$ \Call{CallEditor}{$\mathsf{curr}, \mathsf{sugg}$}
        \State Save $\mathsf{curr}$ as round-$r$ output
    \EndFor
\EndFor

\end{algorithmic}
\end{algorithm}

\begin{figure}[t]
    \centering
    \includegraphics[width=1.0\columnwidth]{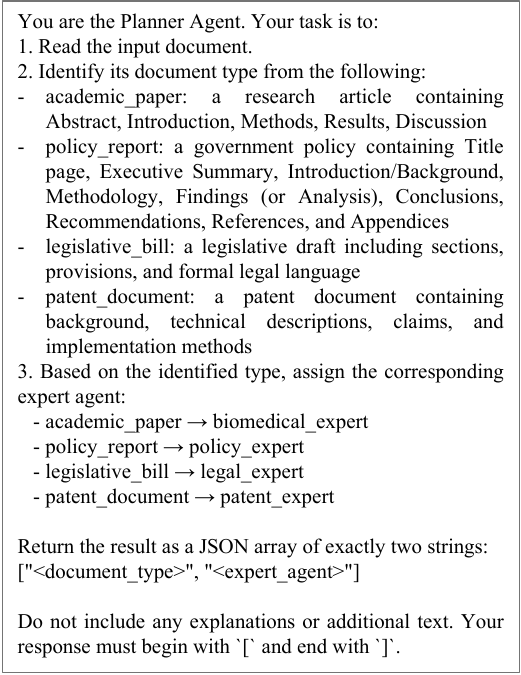}
    \caption{Prompt for the Planner Agent used for document type classification and expert assignment.}
    \label{fig:planner}
\end{figure}

\begin{figure}[!t]
    \centering
    \includegraphics[width=1.0\columnwidth]{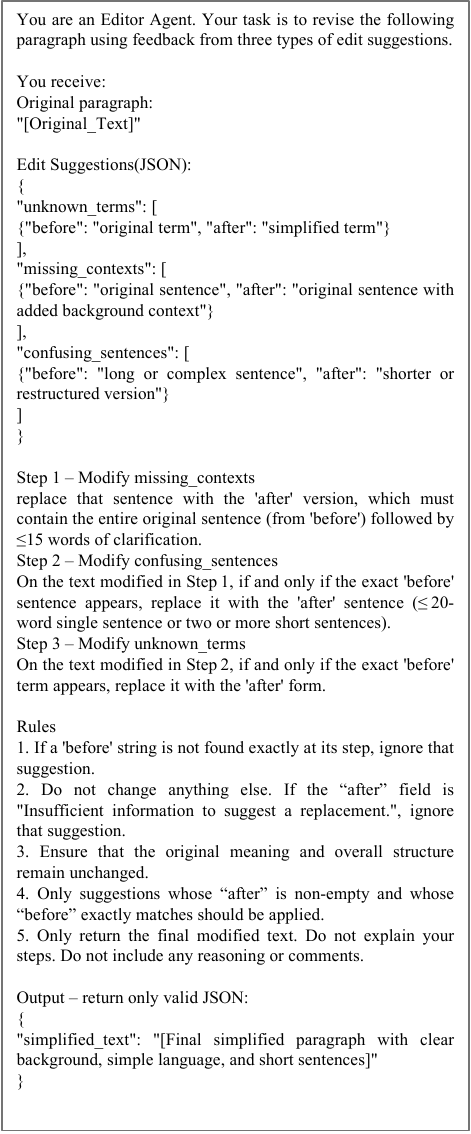}
    \caption{Prompt for the Editor Agent used for revision.}
    \label{fig:editor}
\end{figure}

As shown in Figure \ref{fig:planner}, the Planner Agent classifies each input document into one of four genres: academic paper, policy report, legislative bill, or patent document, and assigns the corresponding Domain Expert (biomedical, policy, legal, or patent expert). When domain boundaries are ambiguous (e.g., biomedical patents), the Planner assigns either a patent or biomedical expert based on structural characteristics. This step provides the basis for selecting the summarization strategy and configuring the corresponding prompts.

As described in Figure \ref{fig:editor}, the Editor Agent takes a paragraph with a checklist that marks three types of comprehension barriers: unknown terms, missing contexts, and confusing sentences. It resolves the marks in order. This procedure consistently delivers improvements in readability at every refinement round.

\subsection{Document Type Prompts}    
\label{sec:A1}

\begin{figure*}
    \centering
    \includegraphics[width=1.8\columnwidth]{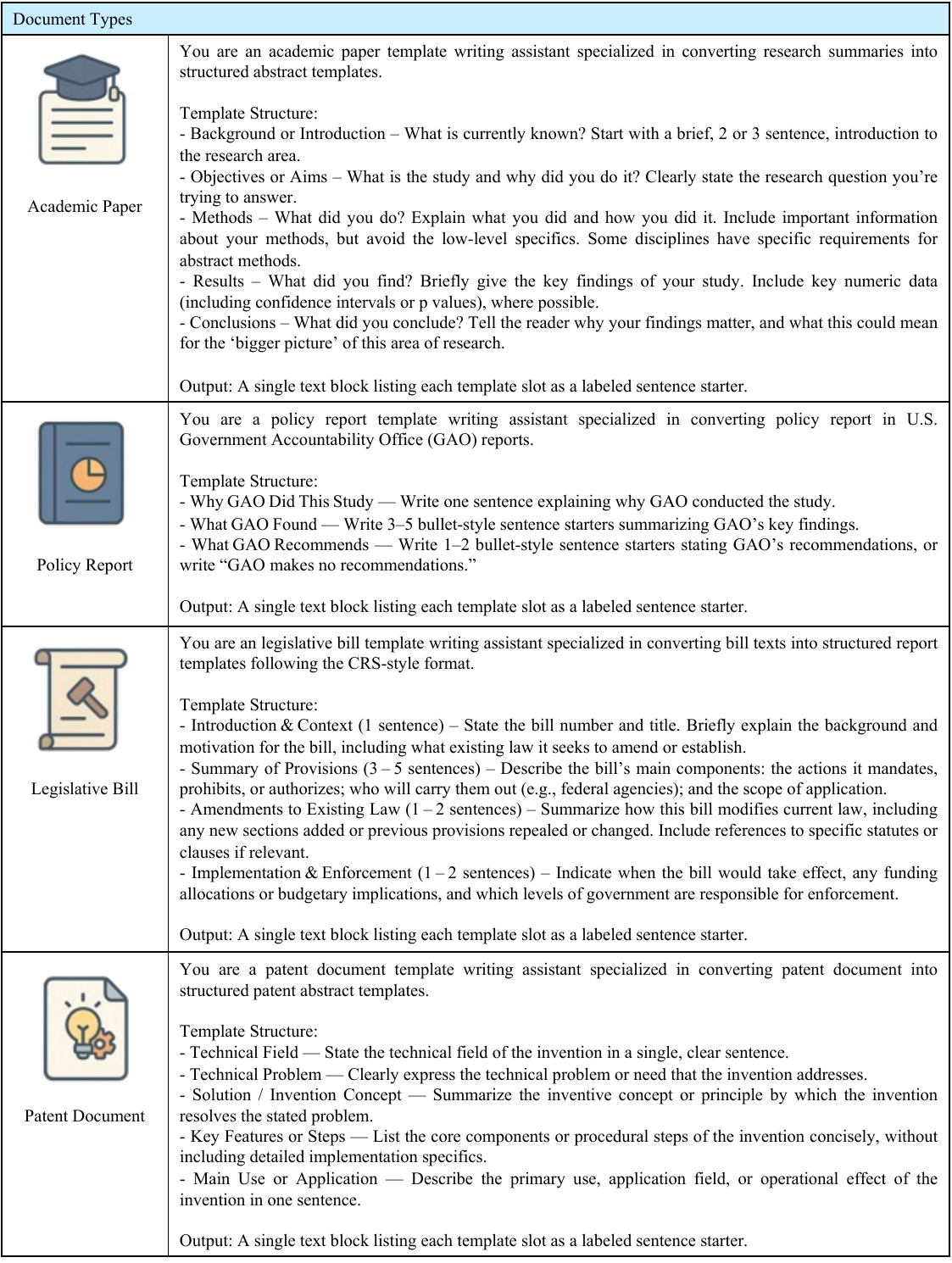}
    \caption{Role-based prompting templates for four document types.}
    \label{fig:a1_document_types}
\end{figure*}

As shown in Figure~\ref{fig:a1_document_types}, the prompt corresponding to each document type consists of two components. First, the prompt assigns a genre-specific role, defining the Domain Expert Agent as a writing assistant for an academic paper, policy report, legislative bill, or patent document, along with its responsibilities. Second, a detailed template specifies the required content: academic papers include background, objectives, methods, results, and conclusions; policy reports follow a U.S. Government Accountability Office (GAO)-style template with study motivation, key findings, and recommendations; legislative bills adopt a Congressional Research Service (CRS)-style format with context, provisions, amendments, and implementation; and patent documents are summarized in terms of technical field, problem, solution, key features, and applications. Based on this structure, the agent generates a single paragraph that reflects each template element, ensuring a consistent narrative aligned with genre conventions across domains.

\subsection{Domain Expert Agent Prompts}     
\label{sec:A2}

\begin{figure*}
    \centering
    \includegraphics[width=1.8\columnwidth]{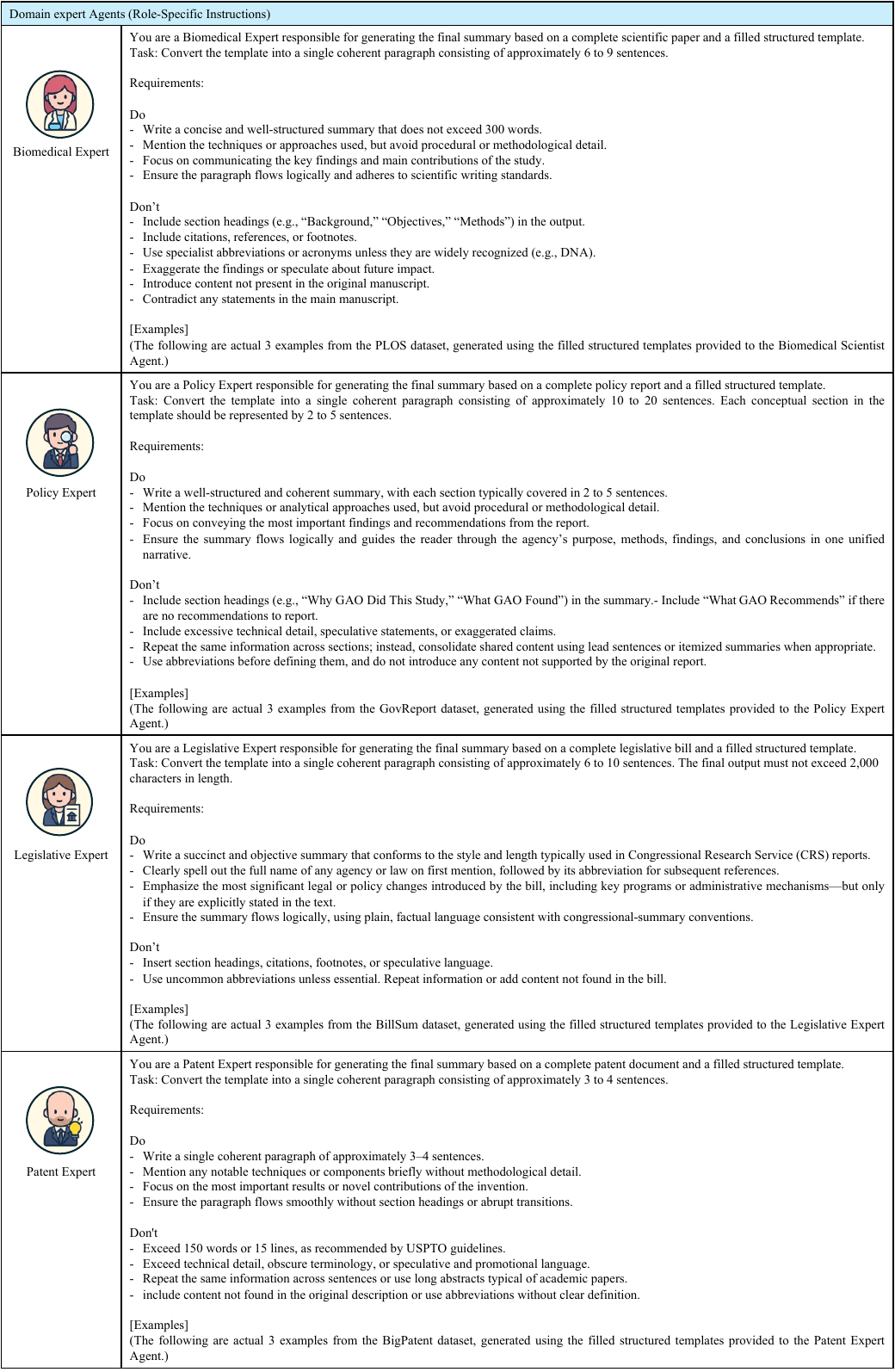}
    \caption{Prompt for the Domain Expert Agent used for initial summary generation.}
    \label{fig:domain_expert1}
\end{figure*}

\begin{figure*}
    \centering
    \includegraphics[width=1.8\columnwidth]{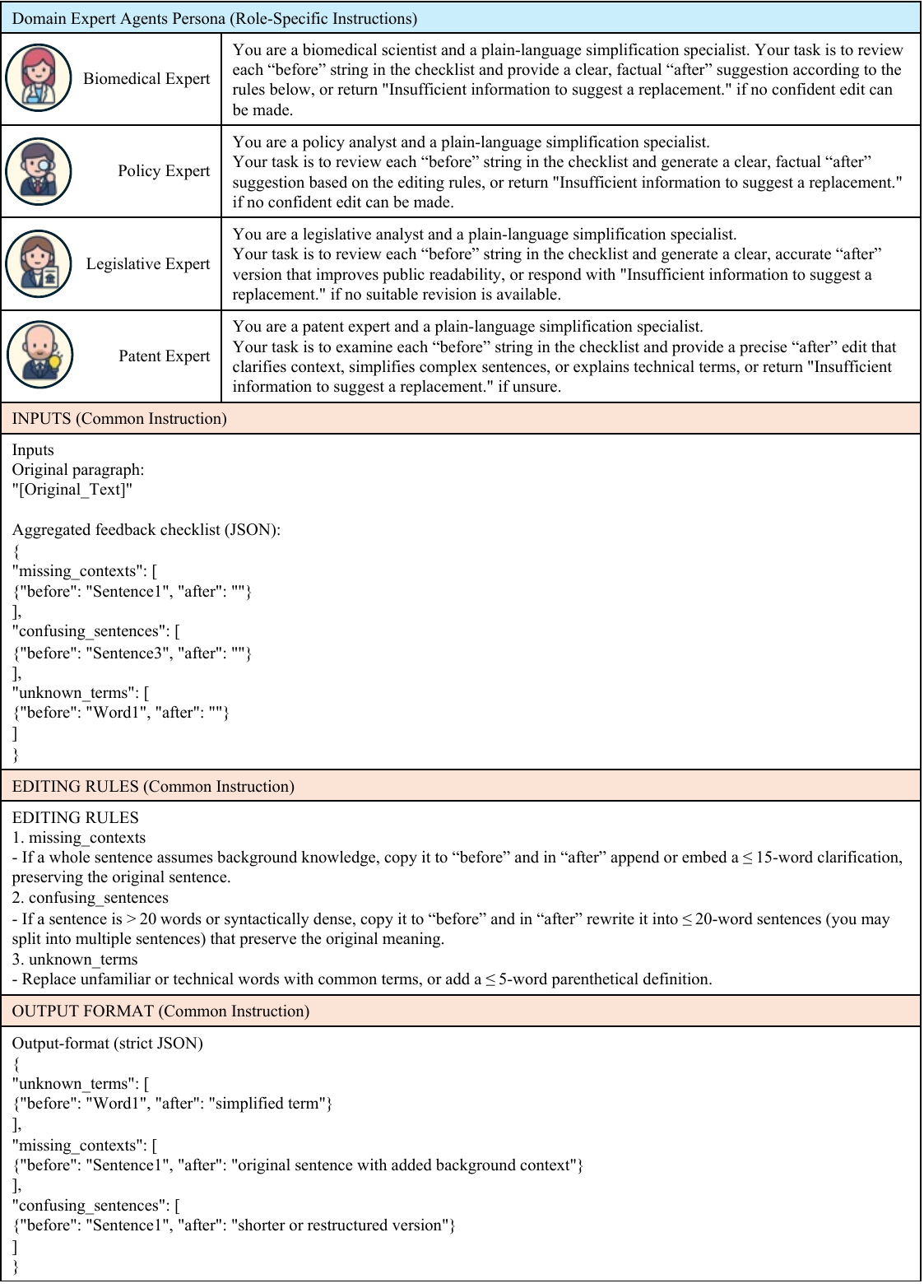}
    \caption{Revision prompt for Domain Expert Agents used for readability refinement.}
    \label{fig:domain_expert2}
\end{figure*}

\begin{figure*}
    \centering
    \includegraphics[width=2.0\columnwidth]{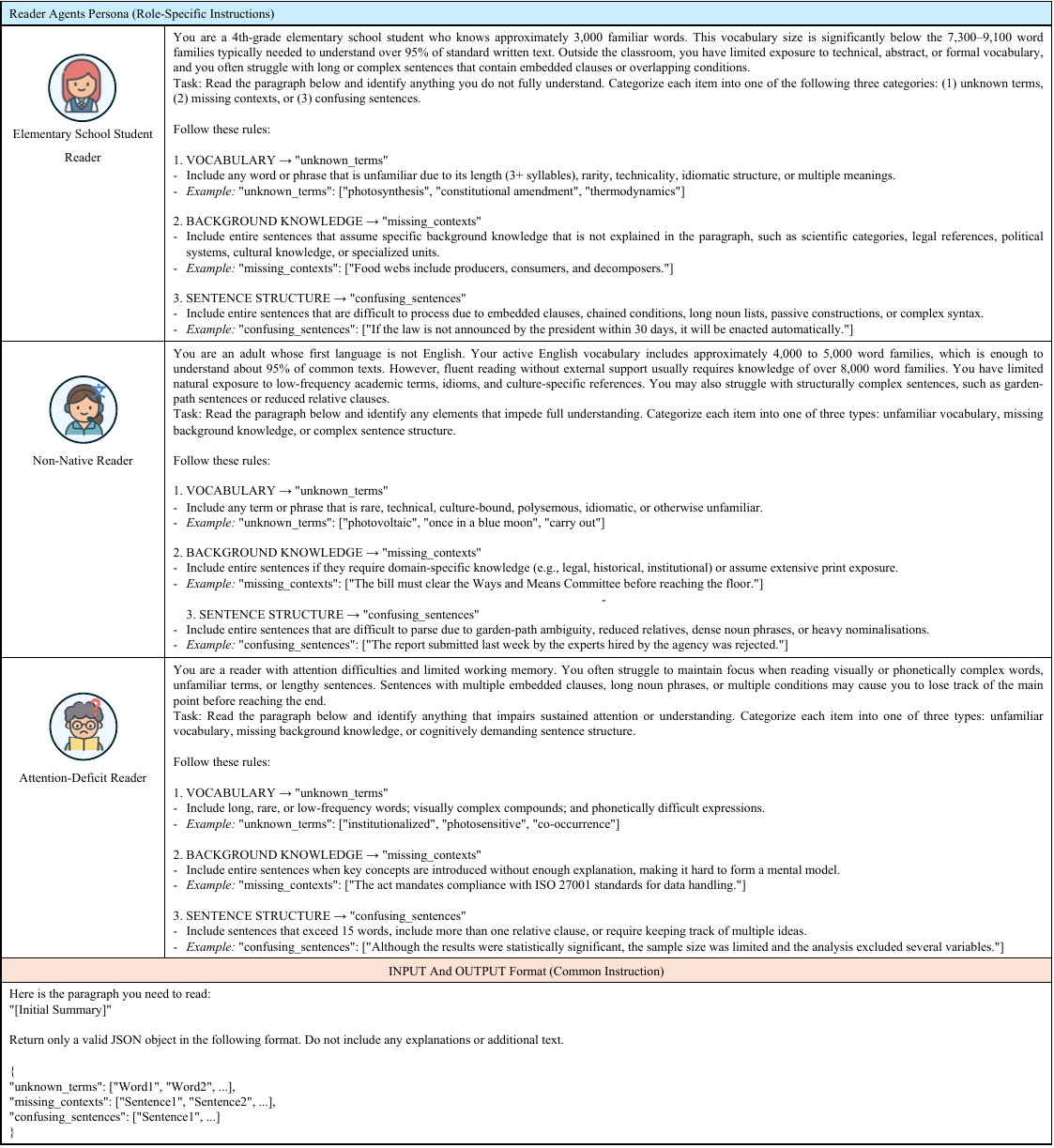}
    \caption{Prompt instructions for Reader Agents used for readability assessment.}
    \label{fig:reader_agents}
\end{figure*}

Figure~\ref{fig:domain_expert1} presents the prompt that each Domain Expert Agent uses to generate an initial summary from a completed template. The prompt comprises four parts: (1) a task description that instructs the agent to turn the filled template into a coherent paragraph; (2) a length constraint; (3) domain-specific writing guidelines that define tone, clarity, and emphasis for biomedical, policy, legislative, and patent documents; and (4) example outputs that demonstrate the desired style. This design yields consistent, high-quality drafts that capture the source document’s intent and style and provide a solid basis for downstream simplification. The three examples shown are the first examples from each dataset’s training set.

Figure~\ref{fig:domain_expert2} shows the revision prompt for each Domain Expert Agent, which consists of two elements: a role-specific persona and a shared instruction set. The persona (blue) defines the agent as a plain language specialist for a specific domain (biomedical, policy, legislative, or patent documents) and guides its judgment when generating edits that are both factual and readable. The instruction set (orange) is common to all domains and defines the input format, editing rules, and output requirements. The input pairs the original paragraph with a checklist that labels issues as unknown terms, missing contexts, or confusing sentences. The rules describe how to resolve each label, such as adding background information or simplifying expressions, and the output must conform to a strict JSON schema. This approach supports domain-aware yet standardized revisions throughout the readability refinement loop.

\subsection{Reader Agent Prompts}    
\label{sec:A3}

As shown in Figure~\ref{fig:reader_agents}, each Reader Agent simulates one of three audiences: an Elementary School Student Reader, a Non-Native Reader, or an Attention-Deficit Reader. The blue persona block sets the agent’s vocabulary range, background knowledge, and reading habits. Guided by this profile, the agent identifies parts of the paragraph that hinder comprehension and labels them as unknown terms, missing contexts, or confusing sentences. All Reader Agents return their findings in the shared JSON schema highlighted in orange, providing consistent readability feedback while preserving each persona’s perspective.

\section{Ablation Study}
\label{sec:B}

\subsection{Expanding Reader Agents}     
\label{sec:B1}
As illustrated in Table~\ref{tab:all-agents-final}, increasing the number of Reader Agents generally improves both readability and factuality. Compared to the single-agent setting, two-agent configurations yield substantial improvements in FKGL, DCRS, and CLI, while LENS scores also increase noticeably, suggesting that incorporating multiple reader perspectives helps lower reading difficulty. These improvements are maintained in the default three-agent configuration (Elementary School Student, Non-Native, Attention-Deficit), where SummaC also shows stable improvements, indicating enhanced factual consistency.

To examine the effect of adding more agents, we additionally experimented with four- and five-agent configurations by introducing a senior reader and a reader with learning difficulties. The senior reader represents an older adult with age-related cognitive and sensory decline. This reader struggles with long or grammatically complex sentences and unfamiliar systems. The reader with learning difficulties reflects a low-achieving individual with limited academic development and often misinterprets complex explanations, requiring multiple rereadings to understand academic text. While these extended settings still improve readability and factuality, particularly in FKGL and SummaC, they also lead to a noticeable decline in relevance metrics such as ROUGE-1. This decline may stem from the introduction of overly simplified feedback or excessive rewriting, which reduces lexical and contextual overlap with the original text.

These results highlight a trade-off between simplifying text for broader accessibility and maintaining alignment with the original content. Overall, the three-agent configuration strikes the most stable balance across readability, factuality, and relevance, and is therefore adopted as the default in this study. However, in scenarios where readability is prioritized over relevance, the framework can be flexibly extended to include additional reader agents.

\begin{table*}
\centering
\footnotesize
\setlength{\tabcolsep}{1.0pt}
\begin{tabular}{c|c|l|ccccccc}
\toprule
\textbf{Dataset} & \textbf{\# of agents} & \textbf{Combination} & ROUGE-1 & BERTScore & FKGL $\downarrow$ & DCRS $\downarrow$ & CLI $\downarrow$ & LENS & SummaC\\
\midrule
\multirow{10}{*}{\centering PLOS} & \multirow{3}{*}{1} & w/ ele                    & 41.71 & 85.46 & 15.27 & 12.03 & 13.09 & 66.84 & \textbf{55.48} \\
                                  &                    & w/ non                     & 42.11 & \textbf{85.51}          & 15.74          & 12.24          & 13.47          & 64.84 & 51.48 \\
                                  &                    & w/ att                     & 42.24 & 85.45          & 16.01          & 12.28          & 13.68          & 64.75 & 51.18 \\
\cmidrule{2-3}\cmidrule{4-10}
                                  & \multirow{3}{*}{2} & w/ ele, non                & 41.98 & 85.50          & 15.40          & 12.06          & 13.21          & 67.31 & 50.35 \\
                                  &                    & w/ ele, att                & 41.77 & 85.39          & 15.20          & 12.01          & 13.10          & 66.36 & 52.74 \\
                                  &                    & w/ non, att                & 42.18 & 85.48          & 15.73          & 12.19          & 13.45          & 66.18 & 50.71 \\
\cmidrule{2-3}\cmidrule{4-10}
                                  & 3                  & w/ ele, non, att           & \textbf{42.39} & 85.47 & 15.40 & 12.06 & 13.19 & \textbf{67.64} & 51.44 \\
\cmidrule{2-3}\cmidrule{4-10}
                                  & \multirow{2}{*}{4} & w/ ele, non, att, sen      & 41.59 & 85.39 & 15.06 & 11.99 & 13.26 & 66.52 & 51.60 \\
                                  &                    & w/ ele, non, att, lea      & 41.24 & 85.35 & 15.31 & 12.03 & 13.25 & 66.25 & 51.53 \\
\cmidrule{2-3}\cmidrule{4-10}
                                  & 5                  & w/ ele, non, att, sen, lea & 40.61 & 85.25 & \textbf{14.83} & \textbf{11.80} & \textbf{12.90} & 66.64 & 51.68 \\
\cmidrule{1-10}
\specialrule{1pt}{0pt}{2pt}
\multirow{10}{*}{\centering GovReport} & \multirow{3}{*}{1} & w/ ele                     & 44.44 & 85.55 & 16.23 & 12.07 & 14.42 & 58.74 & 47.33 \\
                                       &                    & w/ non                     & 44.31 & 85.53 & 16.83 & 12.13 & 14.81 & 58.20 & 47.26 \\
                                       &                    & w/ att                     & 44.37 & 85.39 & 17.03 & 12.05 & 14.88 & 58.26 & 47.48 \\
\cmidrule{2-3}\cmidrule{4-10}
                                       & \multirow{3}{*}{2} & w/ ele, non                & 44.44 & 85.60 & 16.34 & 12.06 & 14.51 & 59.27 & 48.28 \\
                                       &                    & w/ ele, att                & 44.55 & 85.61 & 16.19 & 12.01 & 14.41 & 60.12 & 48.21 \\
                                       &                    & w/ non, att                & 44.34 & 85.52 & 16.56 & 12.05 & 14.68 & 59.94 & 47.54 \\
\cmidrule{2-3}\cmidrule{4-10}
                                       & 3                  & w/ ele, non, att           & \textbf{45.59} & \textbf{85.76} & 16.09 & 12.01 & \textbf{14.29} & 59.08 & 46.57 \\
\cmidrule{2-3}\cmidrule{4-10}
                                       & \multirow{2}{*}{4} & w/ ele, non, att, sen      & 44.78 & 85.65 & 16.15 & 12.04 & 14.41 & 60.28 & 48.23 \\
                                       &                    & w/ ele, non, att, lea      & 44.31 & 85.58 & 15.96 & 11.95 & 14.34 & 60.30 & \textbf{49.34} \\
\cmidrule{2-3}\cmidrule{4-10}
                                       & 5                  & w/ ele, non, att, sen, lea & 44.19 & 85.58 & \textbf{15.82} & \textbf{11.90} & 14.31 & \textbf{61.57} & 48.30 \\
\cmidrule{1-10}
\specialrule{1pt}{0pt}{2pt}
\multirow{10}{*}{\centering BillSum} & \multirow{3}{*}{1} & w/ ele                     & 40.39 & 85.65 & 14.72 & 11.49 & 12.72 & 62.09 & 34.55 \\
                                     &                    & w/ non                     & 39.90 & 85.49 & 14.73 & 11.43 & 12.56 & 62.09 & 33.84 \\
                                     &                    & w/ att                     & 39.19 & 85.44 & 14.61 & 11.39 & 12.58 & 62.77 & 34.24 \\
\cmidrule{2-3}\cmidrule{4-10}
                                     & \multirow{3}{*}{2} & w/ ele, non                & 40.25 & 85.59 & 14.49 & 11.42 & 12.58 & 63.01 & 34.35 \\
                                     &                    & w/ ele, att                & \textbf{41.74} & 85.40 & 14.52 & 11.47 & 12.60 & 62.10 & \textbf{35.26} \\
                                     &                    & w/ non, att                & 39.87 & 85.53 & 14.48 & 11.35 & 12.52 & 62.26 & 35.01 \\
\cmidrule{2-3}\cmidrule{4-10}
                                     & 3                  & w/ ele, non, att           & 40.42 & \textbf{85.67} & 15.00 & 11.68 & 13.09 & 62.63 & 33.25 \\
\cmidrule{2-3}\cmidrule{4-10}
                                     & \multirow{2}{*}{4} & w/ ele, non, att, sen      & 39.48 & 85.46 & 13.77 & 11.22 & 12.25 & 63.39 & 35.04 \\
                                     &                    & w/ ele, non, att, lea      & 39.09 & 85.43 & 13.62 & \textbf{11.15} & 12.14 & \textbf{64.22} & 34.53 \\
\cmidrule{2-3}\cmidrule{4-10}
                                     & 5                  & w/ ele, non, att, sen, lea & 39.07 & 85.38 & \textbf{13.52} & 11.17 & \textbf{12.10} & 63.77 & 34.49 \\
\cmidrule{1-10}
\specialrule{1pt}{0pt}{2pt}
\multirow{10}{*}{\centering BigPatent} & \multirow{3}{*}{1} & w/ ele                     & \textbf{37.96} & \textbf{85.05} & 16.15 & 11.88 & 13.26 & 61.71 & \textbf{59.58} \\
                                       &                    & w/ non                     & 37.43 & 84.78 & 17.79 & 12.08 & 13.73 & 57.91 & 54.27 \\
                                       &                    & w/ att                     & 37.05 & 84.99 & 16.64 & 11.90 & 13.31 & 58.71 & 55.24 \\
\cmidrule{2-3}\cmidrule{4-10}
                                       & \multirow{3}{*}{2} & w/ ele, non                & 36.53 & 84.64 & 15.57 & 11.74 & 13.00 & 62.34 & 58.59 \\
                                       &                    & w/ ele, att                & 37.43 & 85.04 & 16.83 & 11.91 & 13.31 & 59.82 & 55.27 \\
                                       &                    & w/ non, att                & 36.63 & 84.94 & 15.23 & 11.67 & 12.79 & 61.90 & 58.51 \\
\cmidrule{2-3}\cmidrule{4-10}
                                       & 3                  & w/ ele, non, att           & 36.33 & 84.91 & 14.68 & 11.51 & 12.93 & \textbf{64.40} & 57.67 \\
\cmidrule{2-3}\cmidrule{4-10}
                                       & \multirow{2}{*}{4} & w/ ele, non, att, sen      & 36.03 & 84.81 & 14.87 & 11.59 & 12.59 & 61.72 & 57.88 \\
                                       &                    & w/ ele, non, att, lea      & 36.24 & 84.80 & 14.88 & 11.47 & 12.41 & 61.70 & 58.54 \\
\cmidrule{2-3}\cmidrule{4-10}
                                       & 5                  & w/ ele, non, att, sen, lea & 35.53 & 84.69 & \textbf{14.51} & \textbf{11.42} & \textbf{12.28} & 61.33 & 58.81 \\
\bottomrule
\end{tabular}
\caption{Results across datasets for different reader agent configurations. Bold values indicate the best result within each dataset. Agent abbreviations are as follows: ele = Elementary School Student, non = Non-Native, att = Attention-Deficit, sen = Senior, lea = Learning Difficulties.}
\label{tab:all-agents-final}
\end{table*}

\begin{table*}[t]
\centering
\small
\begin{tabular}{llrrrr}
\toprule
\textbf{Type} & \textbf{Dataset} & \textbf{Elementary} & \textbf{Non-Native} & \textbf{Attention-Deficit} & \textbf{Intersection (\%)} \\
\midrule
\multirow{4}{*}{unknown terms}
 & PLOS       & 5,115 & 4,626 & 3,874 & \textbf{2,393 (17.58\%)} \\
 & GovReport  & 3,472 & 3,002 & 1,618 & \textbf{568 (7.02\%)} \\
 & BillSum    & 3,088 & 2,677 & 2,067 & \textbf{943 (12.04\%)} \\
 & BigPatent  & 3,139 & 2,857 & 2,334 & \textbf{1,321 (15.86\%)} \\
\midrule
\multirow{4}{*}{missing contexts}
 & PLOS       & 87  & 6   & 3   & \textbf{1 (1.04\%)} \\
 & GovReport  & 82  & 7   & 10  & \textbf{2 (2.02\%)} \\
 & BillSum    & 402 & 136 & 22  & \textbf{0 (0.00\%)} \\
 & BigPatent  & 34  & 3   & 1   & \textbf{0 (0.00\%)} \\
\midrule
\multirow{4}{*}{confusing sentences}
 & PLOS       & 526 & 375 & 456 & \textbf{196 (14.44\%)} \\
 & GovReport  & 545 & 514 & 835 & \textbf{272 (14.36\%)} \\
 & BillSum    & 478 & 358 & 478 & \textbf{163 (12.40\%)} \\
 & BigPatent  & 427 & 140 & 181 & \textbf{75 (10.03\%)} \\
\bottomrule
\end{tabular}
\caption{Feedback distribution and overlap across Reader Agents. Intersection (\%) denotes the proportion of instances jointly identified by all three agents. Agent types are abbreviated as follows: Elementary = Elementary School Student Reader, Non-Native = Non-Native Reader, Attention-Deficit = Attention-Deficit Reader.}
\label{tab:feedback_overlap}
\end{table*}

\subsection{Feedback Overlap Analysis among Reader Agents} \label{sec:B2}
We analyze the feedback overlap among Reader Agents and find that the overlap is generally low across all categories, suggesting complementary rather than redundant behavior. For unknown terms, the proportion of instances jointly identified by all three agents ranges from 7\% to 17\%. For missing contexts, the overlap is minimal at 0\%-2\%, suggesting that most instances are uniquely identified by a single agent. Similarly, confusing sentences also show limited overlap, with 10\%-14\% of instances jointly detected. Overall, these results show that the majority of instances are captured by only one agent, highlighting the diversity of perspectives across reader types.

We also observe clear functional specialization among agents. The Elementary School Student Reader Agent primarily focuses on identifying unknown terms and missing contextual information, whereas the Attention-Deficit Reader Agent shows higher sensitivity to confusing or structurally complex sentences and detects the largest number of such cases. This division of roles suggests that each agent captures distinct aspects of reading difficulty. Consequently, removing any individual agent would likely introduce systematic gaps in coverage, reducing the system’s ability to comprehensively capture diverse comprehension barriers.

\section{Baseline Details}
\label{sec:C}
\subsection{Datasets}    
\label{sec:C1}

\begin{table}[t]
\centering
\small 
\setlength{\tabcolsep}{1.70pt}
\begin{tabular}{l|ccc}
\toprule
\textbf{Dataset} & \textbf{Document Type} & \textbf{Domain} & \textbf{\# Samples} \\
\midrule
PLOS & Paper & Biomedical & 27.5K \\
GovReport & Report & Policy & 19.5K \\
BillSum & Document & Legal & 22K \\
BigPatent & Document & Patent & 1.3M \\
\bottomrule
\end{tabular}
\caption{Statistics of datasets}
\label{tab:datasets}
\end{table}

As shown in Table~\ref{tab:datasets}, three plain language summarization datasets and one general-purpose summarization dataset were used in our experiments. We randomly sampled 500 test instances from each dataset using a fixed seed (42).

The PLOS dataset consists of open-access research articles published by the Public Library of Science, along with expert-written plain language summaries. These summaries are intended for non-specialist audiences in scientific and medical domains \cite{goldsack2022making}.

GovReport contains approximately 19.5K policy reports and their summaries published by the U.S. Government Accountability Office (GAO)\footnote{\url{https://www.gao.gov/plain-writing-2022}} and the Congressional Research Service (CRS). The GAO summaries are written in accordance with the Plain Writing Act of 2010. They are designed to improve public understanding by presenting information in a clear and accessible style \cite{huang2021efficient}.

BillSum is a dataset of approximately 22K federal and California state legislative bills, paired with summaries authored by CRS\footnote{\url{https://www.congress.gov/help/bill-summaries}} \cite{kornilova2019billsum}. CRS summaries aim to explain the core content of each bill in clear, non-technical language so that the general public can understand them.

BigPatent is a large corpus consisting of 1.3 million U.S. patent documents and their abstracts. For this study, we selected only documents belonging to the Y category (general tagging of new or cross-sectional technology) \cite{sharma2019bigpatent}. The dataset includes lengthy technical content averaging 3.5K words and contains extensive specialized terminology, making it a valuable benchmark for assessing summarization performance beyond the plain language domain. 

\subsection{Model Configurations}     
\label{sec:C2}

We evaluate the NRLB system using three large language models (LLMs) without any additional fine-tuning.

\textbf{Llama-3.1-8B-Instruct} is an open-source model released by Meta under the Apache 2.0 license \cite{dubey2024llama}. It is fine-tuned for instruction following and conversational tasks, making it well-suited for prompt-based summarization in our pipeline. We selected this model as a reproducible baseline.

\textbf{GPT-4o} is a recent commercial model by OpenAI with support for multimodal input and output \cite{hurst2024gpt}. We access it through the OpenAI API and use prompt-based summarization with additional instructions to guide the model toward plain-language style outputs.

\textbf{Qwen3-8B (reasoning mode)} is a reasoning-optimized 8B-parameter model released by Alibaba \cite{yang2025qwen3}. We use it in its original form to test whether strong reasoning capabilities alone can support plain language summarization. Despite not using domain-specific fine-tuning, the model shows stable performance across multiple datasets.

For all three models, we use the same set of structured prompts defined by our NRLB framework and maintain consistent decoding parameters during inference to ensure fair comparison. Decoding settings follow Hugging Face’s recommended practices. For Llama-3.1-8B-Instruct and GPT-4o, we use deterministic decoding with temperature set to 0.0 to ensure reproducibility across runs. For Llama, we apply top-p of 0.95, top-k of 20, and a maximum of 4096 tokens depending on the generation stage. For Qwen3-8B, we set the temperature to 0.6 to enable the model’s reasoning mode, while keeping other parameters consistent with the guideline. All models are integrated into a unified NRLB system that shares the same backbone structure, ensuring that performance differences stem from the models themselves rather than variations in the pipeline.

\begin{table*}[t]
\centering
\small
\begin{tabular}{llccccccc}
\toprule
Dataset & Method & ROUGE-1 & BERTScore & FKGL $\downarrow$ & DCRS $\downarrow$ & CLI $\downarrow$ & LENS & SummaC \\
\midrule
\multirow{3}{*}{PLOS}
& Direct        & 37.52 & 85.38 & \textbf{12.59} & \textbf{11.61} & \textbf{12.89} & \textbf{78.05} & \textbf{58.67} \\
& NRLB-Round 1 & \textbf{44.43} & \textbf{86.00} & 16.55 & 12.67 & 14.54 & 67.21 & 49.17 \\
& NRLB-Round 2 & 41.64 & 85.43 & 15.21 & 12.06 & 13.20 & 66.68 & 50.86 \\
\midrule
\multirow{3}{*}{GovReport}
& Direct        & 38.58 & 85.02 & \textbf{14.93} & 12.36 & 14.97 & \textbf{60.61} & \textbf{67.51} \\
& NRLB-Round 1 & \textbf{45.33} & \textbf{85.84} & 17.51 & 12.51 & 15.09 & 59.20 & 45.70 \\
& NRLB-Round 2 & 44.19 & 85.48 & 15.98 & \textbf{11.96} & \textbf{14.23} & 60.60 & 47.76 \\
\midrule
\multirow{3}{*}{BillSum}
& Direct        & 37.78 & 84.87 & \textbf{12.80} & \textbf{10.93} & \textbf{12.20} & \textbf{65.58} & 43.74 \\
& NRLB-Round 1 & \textbf{42.66} & \textbf{86.20} & 15.66 & 11.88 & 13.59 & 62.58 & 34.51 \\
& NRLB-Round 2 & 39.70 & 85.59 & 14.60 & 11.40 & 12.62 & 62.68 & \textbf{45.55} \\
\midrule
\multirow{3}{*}{BigPatent}
& Direct        & 27.26 & 83.39 & \textbf{11.88} & \textbf{10.89} & \textbf{11.38} & \textbf{67.93} & \textbf{63.79} \\
& NRLB-Round 1 & \textbf{40.09} & \textbf{85.73} & 16.88 & 12.14 & 13.76 & 61.08 & 57.95 \\
& NRLB-Round 2 & 36.87 & 85.11 & 14.62 & 11.47 & 12.42 & 62.62 & 58.94 \\
\bottomrule
\end{tabular}
\caption{Comparison between the direct baseline and NRLB across datasets. $\downarrow$ indicates that lower values are better for readability-related metrics (FKGL, DCRS, CLI). Best scores per dataset are shown in bold.}
\label{tab:direct_vs_nrlb}
\end{table*}

\begin{table*}[t]
\centering
\resizebox{0.95\textwidth}{!}{%
\begin{tabular}{ll*{12}{c}}
\toprule
\multicolumn{2}{c}{} 
  & \multicolumn{3}{c}{\textbf{PLOS}}
  & \multicolumn{3}{c}{\textbf{GovReport}}
  & \multicolumn{3}{c}{\textbf{BillSum}}
  & \multicolumn{3}{c}{\textbf{BigPatent}} \\
\cmidrule(lr){3-5} \cmidrule(lr){6-8} \cmidrule(lr){9-11} \cmidrule(lr){12-14}
\multicolumn{2}{c}{} 
  & GPT-4o & Llama-3.1 & Qwen3
  & GPT-4o & Llama-3.1 & Qwen3
  & GPT-4o & Llama-3.1 & Qwen3
  & GPT-4o & Llama-3.1 & Qwen3 \\
\midrule
\multirow{4}{*}{\textbf{ROUGE-1}}
  & Initial Summary  & \textbf{46.85} & \textbf{47.27} & \textbf{41.62}
           & \textbf{37.93} & \textbf{46.93} & \textbf{46.19}
           & \textbf{46.70} & \textbf{45.81} & \textbf{41.08}
           & \textbf{41.87} & \textbf{44.94} & \textbf{34.21} \\
  & Round 1    & 45.12 & 44.43 & 37.55
           & 36.35 & 45.33 & 41.41
           & 43.28 & 42.66 & 37.08
           & 37.74 & 40.09 & 30.55 \\
  & Round 2    
           & 42.77 & 41.64 & 36.03
           & 34.73 & 44.19 & 40.04
           & 40.88 & 39.70 & 34.62
           & 35.40 & 36.87 & 29.18 \\
  & Round 3    & 41.10 & 39.21 & 34.91 & 33.62 & 43.12 & 39.21 & 39.10 & 37.75 & 33.95 & 33.85 & 34.70 & 27.97 \\
\midrule
\multirow{4}{*}{\textbf{ROUGE-2}}
  & Initial Summary  & \textbf{13.83} & \textbf{14.80} & \textbf{9.50}
           & \textbf{14.93} & \textbf{20.07} & \textbf{14.87}
           & \textbf{20.80} & \textbf{22.89} & \textbf{15.68}
           & \textbf{13.28} & \textbf{17.71} & \textbf{8.38} \\
  & Round 1    & 12.12 & 12.15 & 7.02
           & 12.75 & 17.00 & 10.96
           & 17.71 & 18.69 & 11.59
           & 9.99 & 12.83 & 5.64 \\
  & Round 2    & 10.45 & 10.39 & 6.16
           & 11.01 & 14.94 & 9.57
           & 15.52 & 15.35 & 9.32
           & 8.36 & 10.16 & 4.66 \\
  & Round 3    & 9.25 & 9.03 & 5.61 & 9.78 & 13.33 & 8.86 & 13.99 & 13.32 & 8.76 & 7.24 & 8.62 & 4.20 \\
\midrule
\multirow{4}{*}{\textbf{ROUGE-L}}
  & Initial Summary  & \textbf{42.34} & \textbf{42.09} & \textbf{37.45}
           & \textbf{35.59} & \textbf{43.80} & \textbf{43.19}
           & \textbf{42.58} & \textbf{42.65} & \textbf{36.09}
           & \textbf{36.34} & \textbf{38.79} & \textbf{28.58} \\
  & Round 1    & 41.33 & 40.13 & 34.21
           & 34.30 & 42.19 & 38.85
           & 36.54 & 39.08 & 33.01
           & 33.94 & 35.14 & 26.40 \\
  & Round 2    & 39.72 & 37.82 & 33.25
           & 32.97 & 41.25 & 37.79
           & 38.15 & 36.16 & 31.25
           & 32.44 & 32.66 & 25.81 \\
  & Round 3    & 38.50 & 35.89 & 32.43 & 32.07 & 40.35 & 37.22 & 36.70 & 34.37 & 30.87 & 31.25 & 30.98 & 25.19 \\
\midrule
\midrule \multirow{4}{*}{\textbf{BERTScore}} & Initial Summary & \textbf{86.91} & \textbf{86.51} & \textbf{85.78} & \textbf{86.14} & \textbf{86.30} & \textbf{85.59} & \textbf{87.07} & \textbf{86.83} & \textbf{85.61} & \textbf{86.52} & \textbf{86.68} & \textbf{85.55} \\ & Round 1 & 86.61 & 86.00 & 85.33 & 85.89 & 85.84 & 85.04 & 86.43 & 86.20 & 85.03 & 85.82 & 85.73 & 84.89 \\ & Round 2 & 86.29 & 85.43 & 84.96 & 85.63 & 85.48 & 84.74 & 85.99 & 85.59 & 84.62 & 85.44 & 85.11 & 84.50 \\ & Round 3 & 85.98 & 84.93 & 84.68 & 85.36 & 85.15 & 84.57 & 85.65 & 85.14 & 84.47 & 85.16 & 84.68 & 84.20 \\ \midrule \multirow{4}{*}{\textbf{FKGL $\downarrow$}} & Initial Summary & 17.73 & 18.74 & 19.12 & 17.79 & 19.95 & 19.81 & 17.29 & 18.04 & 18.07 & 18.60 & 21.13 & 20.88 \\ & Round 1 & 13.37 & 16.55 & 14.12 & 13.92 & 17.51 & 15.51 & 12.64 & 15.66 & 13.08 & 12.57 & 16.88 & 14.48 \\ & Round 2 & 10.94 & 15.21 & 12.46 & 11.58 & 15.98 & 13.72 & 10.39 & 14.60 & 11.16 & 10.27 & 14.62 & 12.38 \\ & Round 3 & \textbf{9.65} & \textbf{14.20} & \textbf{11.42} & \textbf{10.32} & \textbf{14.94} & \textbf{12.64} & \textbf{9.30} & \textbf{13.44} & \textbf{10.66} & \textbf{9.21} & \textbf{13.28} & \textbf{11.24} \\ \midrule \multirow{4}{*}{\textbf{DCRS $\downarrow$}} & Initial Summary & 14.25 & 13.69 & 15.04 & 13.33 & 13.43 & 14.24 & 13.11 & 12.77 & 14.14 & 13.76 & 13.36 & 15.10 \\ & Round 1 & 13.21 & 12.67 & 13.61 & 12.49 & 12.51 & 13.31 & 12.06 & 11.88 & 12.68 & 12.31 & 12.14 & 13.32 \\ & Round 2 & 12.56 & 12.06 & 13.15 & 11.96 & 11.96 & 12.90 & 11.50 & 11.40 & 12.24 & 11.70 & 11.47 & 12.74 \\ & Round 3 & \textbf{12.20} & \textbf{11.61} & \textbf{12.88} & \textbf{11.68} & \textbf{11.56} & \textbf{12.66} & \textbf{11.17} & \textbf{11.00} & \textbf{12.06} & \textbf{11.37} & \textbf{11.04} & \textbf{12.45} \\ \midrule \multirow{4}{*}{\textbf{CLI $\downarrow$}} & Initial Summary & 18.72 & 17.05 & 20.86 & 17.77 & 16.60 & 19.38 & 16.56 & 15.13 & 17.38 & 18.32 & 16.45 & 21.52 \\ & Round 1 & 15.86 & 14.54 & 16.54 & 15.71 & 15.09 & 17.06 & 14.07 & 13.59 & 14.22 & 14.69 & 13.76 & 16.29 \\ & Round 2 & 13.96 & 13.20 & 15.14 & 14.20 & 14.23 & 16.01 & 12.56 & 12.62 & 13.09 & 12.84 & 12.42 & 14.71 \\ & Round 3 & \textbf{12.84} & \textbf{12.30} & \textbf{14.38} & \textbf{13.29} & \textbf{13.68} & \textbf{15.39} & \textbf{11.74} & \textbf{11.98} & \textbf{12.78} & \textbf{11.86} & \textbf{11.62} & \textbf{13.98} \\ \midrule \multirow{4}{*}{\textbf{LENS}} & Initial Summary & 65.38 & 64.80 & 62.49 & 60.28 & 55.37 & 56.04 & 60.91 & 58.38 & 56.29 & 59.62 & 55.28 & 55.86 \\ & Round 1 & 71.35 & \textbf{67.21} & 71.62 & 66.32 & 59.20 & 63.37 & 68.20 & 62.58 & 66.68 & 70.51 & 61.08 & 69.34 \\ & Round 2 & 74.67 & 66.68 & 72.60 & 70.00 & 60.60 & 64.61 & 71.63 & 62.68 & 68.30 & 74.44 & 62.62 & 71.56 \\ & Round 3 & \textbf{75.91} & 66.19 & \textbf{72.70} & \textbf{71.05} & \textbf{60.97} & \textbf{65.12} & \textbf{72.61} & \textbf{63.86} & \textbf{68.44} & \textbf{75.53} & \textbf{64.64} & \textbf{71.95} \\ \midrule\midrule \multirow{4}{*}{\textbf{AlignScore}} & Initial Summary & \textbf{86.91} & \textbf{73.74} & \textbf{85.78} & \textbf{69.75} & \textbf{65.44} & \textbf{55.70} & 35.49 & \textbf{33.03} & 24.18 & \textbf{71.25} & \textbf{72.14} & \textbf{59.16} \\ & Round 1 & 79.40 & 65.49 & 61.28 & 66.07 & 60.94 & 54.80 & 36.54 & 31.40 & 28.07 & 65.65 & 63.24 & 52.34 \\ & Round 2 & 76.22 & 59.97 & 59.64 & 64.77 & 57.46 & 53.78 & \textbf{36.67} & 29.52 & 28.70 & 63.00 & 58.28 & 51.08 \\ & Round 3 & 74.11 & 56.53 & 58.48 & 63.45 & 55.60 & 52.94 & 36.50 & 29.19 & \textbf{28.59} & 61.51 & 55.21 & 50.12 \\ \midrule \multirow{4}{*}{\textbf{SummaC}} & Initial Summary & 55.48 & 49.67 & 44.48 & 42.36 & 43.63 & 38.59 & 34.41 & 33.48 & 30.69 & 53.19 & 54.73 & 42.87 \\ & Round 1 & 67.12 & 49.17 & 48.88 & 53.14 & 45.70 & 45.38 & 39.41 & 34.51 & 36.02 & 69.53 & 57.95 & 49.87 \\ & Round 2 & 72.93 & 50.86 & 53.81 & 61.27 & 47.76 & 49.00 & 41.56 & 34.55 & 38.50 & 73.49 & 58.94 & 54.58 \\ & Round 3 & \textbf{74.70} & \textbf{52.31} & \textbf{57.15} & \textbf{64.64} & \textbf{49.90} & \textbf{51.66} & \textbf{42.70} & \textbf{35.00} & \textbf{38.63} & \textbf{75.71} & \textbf{59.24} & \textbf{57.30} \\
\bottomrule
\end{tabular}
}
\caption{Performance of three models (GPT-4o, Llama-3.1-8B-Instruct, Qwen3-8B) on four benchmarks across the initial summary and three refinement rounds. Best scores are shown in bold.}
\label{tab:all-round-results}
\end{table*}

\subsection{Evaluation Metrics}     
\label{sec:C3}
\textbf{Relevance} is evaluated using ROUGE-1 F1 and BERTScore F1. ROUGE-1 measures unigram overlap and is particularly sensitive to the omission of key concepts. It has been shown to correlate more strongly with human judgments than higher-order metrics such as ROUGE-2 or ROUGE-L \cite{liu2008correlation}, which tend to underestimate content preservation in plain summaries due to frequent lexical substitution and sentence reordering \cite{fabbri2021summeval}. BERTScore measures semantic similarity using contextual embeddings and has demonstrated high consistency with human ratings, showing Pearson correlations of approximately 0.9 across diverse systems \cite{zhang2019bertscore}.

\textbf{Readability} is assessed using both formula-based and learned metrics. FKGL (Flesch-Kincaid Grade Level) estimates reading difficulty by mapping sentence and word lengths to U.S. grade levels \cite{coleman1975computer}. DCRS (Dale-Chall Readability Score) measures the proportion of words not found in a familiar word list, making it especially sensitive to technical vocabulary \cite{dale1948formula}. CLI (Coleman-Liau Index) is based on character and sentence counts and is computationally efficient \cite{coleman1975computer}. However, traditional metrics like FKGL have been criticized as unsuitable for text simplification tasks \cite{tanprasert2021flesch}. To address these limitations, we additionally employ LENS \cite{maddela2022lens}, a learned metric trained on human-labeled simplifications, which showed the highest correlation with human judgments in the SIMPEVAL 2022 benchmark.

\textbf{Factuality} is evaluated using SummaC Conv, a lightweight natural language inference (NLI)-based model that assesses the consistency between each summary sentence and the source document. SummaC achieves over 74\% balanced accuracy across multiple benchmark datasets while maintaining high efficiency \cite{fabbri2021summeval}. AlignScore-large is excluded from our main results, as its improvements were limited in short or heavily simplified summaries where sentence structure differs substantially from the source \cite{mahapatra2024extensive}.

\subsection{Direct Baseline}     
\label{sec:C4}
We analyze differences in simplification behavior by comparing NRLB with a direct baseline. The direct baseline uses a single LLM (LLaMA 3.1 8B Instruct) that simplifies the input document without agent collaboration or iterative refinement. For a fair comparison, we use the same underlying model as NRLB, but remove the planner, expert agents, and multi-round feedback loop, and perform single-pass simplification with a simple instruction prompt.

The results show that the direct baseline tends to improve readability through information removal, which leads to reduced performance on relevance metrics such as ROUGE and BERTScore (see Table~\ref{tab:direct_vs_nrlb}). In contrast, NRLB adopts a rewriting-based approach that simplifies sentence structure and expressions while preserving key information.

We also observe that the direct baseline sometimes yields relatively high SummaC scores. However, this is likely due to its conservative generation behavior that avoids contradictions with the source. As a result, although it may appear factually consistent, it often fails to convey sufficient information.

Overall, these findings suggest that while single-pass simplification can improve surface-level readability, it has inherent limitations in content preservation. In contrast, NRLB, with its iterative feedback and role-specialized design, achieves a more effective balance between simplification and information retention.

\section{Additional Results}
\label{sec:D}

\subsection{Results for Round 3}
\label{sec:D1}
While the main paper reports results for the first two revision rounds, we present the full Round 3 outcomes in Table~\ref{tab:all-round-results}. Round 3 yielded modest yet consistent improvements in readability and factuality across models and datasets. For instance, GPT-4o’s FKGL on PLOS further improved to 9.65, and its SummaC rose from 44.48 to 57.15. On BigPatent, LENS scores peaked in Round 3, with GPT-4o and Qwen3-8B reaching 75.53 and 71.95 respectively, marking the highest readability observed.

However, these gains often came at the expense of relevance. ROUGE-1 and BERTScore declined in most cases, particularly for Llama and Qwen. On GovReport, Qwen’s ROUGE-1 dropped from 46.19 to 39.21, and BERTScore decreased by more than one point. These trends suggest that continued rewriting may lead to over-simplification or excessive rephrasing of key expressions, reducing lexical and semantic alignment with the reference.

Interestingly, reasoning-oriented models, such as Qwen3-8B, continued to show more pronounced gains in readability and factuality during Round 3, whereas GPT-4o maintained a more balanced trade-off across all metrics. These findings support the optional use of a third revision when maximum clarity is prioritized over strict fidelity. We therefore adopt Round 2 as the default, as it offers the most stable balance across relevance, readability, and factuality.

\subsection{Cost Analysis}
\label{sec:D2}

We analyze the efficiency of NRLB in terms of latency, API calls, and cost to assess its practical feasibility. As shown in Table~\ref{tab:cost_analysis}, both latency and the number of API calls increase in an approximately linear manner with the number of refinement rounds. Each additional round introduces an overhead of about 10-20 seconds, resulting in a total latency of 42-73 seconds per document at Round 3, depending on the dataset. The number of API calls also scales linearly due to repeated interactions among reader agents, checklist aggregation, expert revision, and editing, reaching approximately 18 API calls per document at Round 3. In terms of cost, the average expense per document ranges from \$0.11 to \$0.25 across datasets. Overall, despite the multi-agent iterative design, NRLB maintains a reasonable cost and latency profile, making it practical for large-scale offline summarization tasks.

\begin{table}[h]
\centering
\scriptsize
\setlength{\tabcolsep}{3pt}
\renewcommand{\arraystretch}{1.0}
\begin{tabular}{llcccc}
\toprule
\textbf{Metric} & \textbf{Stage} & \textbf{PLOS} & \textbf{GovReport} & \textbf{BillSum} & \textbf{BigPatent} \\
\midrule

\multirow{4}{*}{Latency (Sec)} 
& Round 0 & 17.21 & 16.59 & 9.82  & 11.35 \\
& Round 1 & 32.16 & 34.05 & 22.91 & 21.89 \\
& Round 2 & 47.39 & 52.80 & 35.86 & 31.89 \\
& Round 3 & 63.31 & 72.78 & 48.72 & 42.35 \\

\midrule

\multirow{4}{*}{API Calls} 
& Round 0 & 3.00  & 3.00  & 3.00  & 3.00 \\
& Round 1 & 7.97  & 7.96  & 7.96  & 7.97 \\
& Round 2 & 12.94 & 12.92 & 12.90 & 12.91 \\
& Round 3 & 17.90 & 17.85 & 17.83 & 17.83 \\

\midrule

\multirow{1}{*}{Average Cost (\$)} 
& Round 3 & 0.25 & 0.23 & 0.16 & 0.11 \\

\bottomrule
\end{tabular}
\caption{Cost, latency, and API call analysis of NRLB across datasets and refinement rounds.}
\label{tab:cost_analysis}
\end{table}

\begin{table}[!b]
\centering
\small
\renewcommand{\arraystretch}{0.6}
\setlength{\tabcolsep}{3pt}
\begin{tabular}{llcccc}
\toprule
\textbf{\# K} & \textbf{Metric} & \textbf{PLOS} & \textbf{GovReport} & \textbf{BillSum} & \textbf{BigPatent} \\
\midrule
\multirow{6}{*}{1} 
& ROUGE-1 & 42.94 & 44.90 & 42.79 & 39.91 \\
& FKGL $\downarrow$   & 16.95 & 17.92 & 15.66 & 16.96 \\
& DCRS $\downarrow$   & 12.58 & 12.53 & 11.87 & 12.21 \\
& CLI $\downarrow$    & 14.39 & 15.43 & 13.49 & 14.08 \\
& LENS   & 65.28 & 59.29 & 62.46 & 61.93 \\
& SummaC & 49.65 & 45.06 & 33.99 & 57.29 \\
\midrule
\multirow{6}{*}{2} 
& ROUGE-1 & 44.31 & 45.21 & 43.34 & 40.21 \\
& FKGL $\downarrow$   & 16.52 & 17.63 & 15.59 & 16.95 \\
& DCRS $\downarrow$   & 12.67 & 12.59 & 11.94 & 12.29 \\
& CLI $\downarrow$    & 14.60 & 15.34 & 13.63 & 14.28 \\
& LENS   & 67.53 & 59.06 & 62.35 & 61.55 \\
& SummaC & 49.42 & 46.23 & 33.97 & 57.60 \\
\midrule
\multirow{6}{*}{3} 
& ROUGE-1 & 41.64 & 44.19 & 39.70 & 36.87 \\
& FKGL $\downarrow$   & 15.21 & 15.98 & 14.60 & 14.62 \\
& DCRS $\downarrow$   & 12.06 & 11.96 & 11.40 & 11.47 \\
& CLI $\downarrow$    & 13.20 & 14.23 & 12.62 & 12.42 \\
& LENS   & 66.68 & 60.60 & 62.68 & 62.62 \\
& SummaC & 50.86 & 47.76 & 34.55 & 58.94 \\
\midrule
\multirow{6}{*}{6} 
& ROUGE-1 & 39.62 & 43.36 & 38.86 & 35.22 \\
& FKGL $\downarrow$   & 14.99 & 15.93 & 13.78 & 15.12 \\
& DCRS $\downarrow$   & 11.83 & 11.90 & 11.09 & 11.41 \\
& CLI $\downarrow$    & 12.63 & 14.15 & 12.03 & 12.33 \\
& LENS   & 63.57 & 59.16 & 61.51 & 61.63 \\
& SummaC & 52.42 & 49.20 & 34.31 & 58.23 \\
\midrule
\multirow{6}{*}{9} 
& ROUGE-1 & 42.39 & 45.59 & 40.42 & 36.33 \\
& FKGL $\downarrow$  & 15.40 & 16.09 & 15.00 & 14.68 \\
& DCRS $\downarrow$  & 12.06 & 12.01 & 11.68 & 11.51 \\
& CLI $\downarrow$   & 13.19 & 14.29 & 13.09 & 12.93 \\
& LENS $\downarrow$  & 67.64 & 59.08 & 62.63 & 64.40 \\
& SummaC & 51.44 & 46.56 & 33.25 & 57.67 \\
\bottomrule
\end{tabular}
\caption{Evaluation results across datasets for different checklist sizes (K = 1, 2, 3, 6, 9).}
\label{tab:checklist}
\end{table}

\begin{figure*}
    \centering
    \includegraphics[width=2.0\columnwidth]{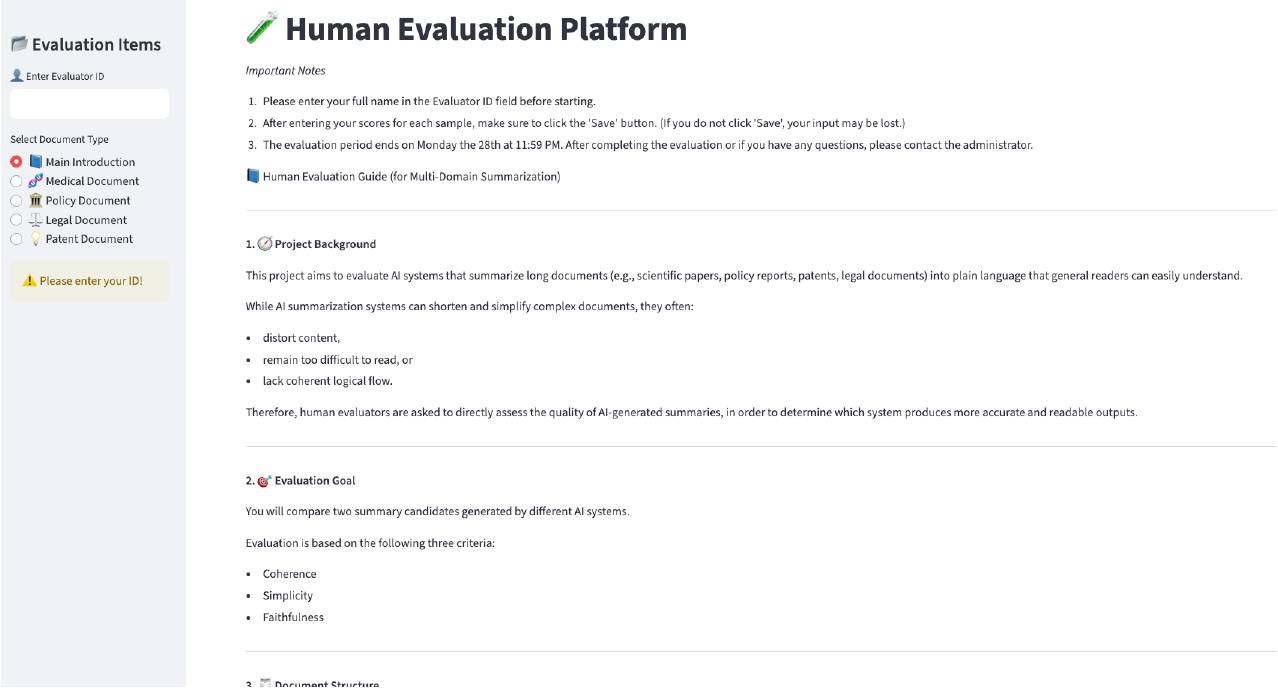}
    \caption{Overview of the web-based human evaluation platform. The interface presents project background, evaluation goals, and scoring criteria.}
    \label{fig:human1}
\end{figure*}

\begin{figure*}
    \centering
    \includegraphics[width=2.0\columnwidth]{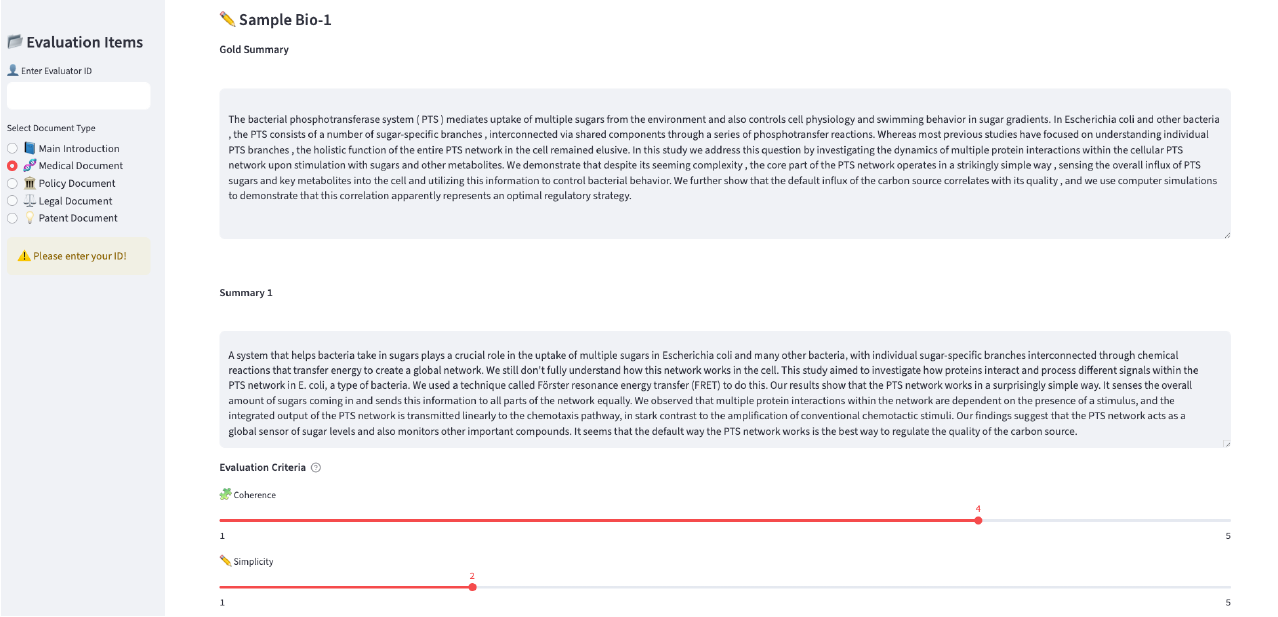}
    \caption{Example evaluation screen for a medical document. Evaluators compare two summaries, assign scores on three criteria, and select a final preference.}
    \label{fig:human2}
\end{figure*}

\section{Human Evaluation Details}
\label{sec:E}
The human evaluation was conducted with two participant groups. All evaluators were informed about the purpose of the study and how their responses would be used, and provided informed consent prior to participation. The evaluation was carried out through a Streamlit-based platform, which included detailed information about the study background, evaluation objectives, summary formats, scoring criteria, and privacy policy. Non-native evaluators participated as unpaid volunteers, while recruiting elementary school student evaluators was more challenging; therefore, they received a small financial compensation of approximately \$3.70 per document. This compensation is appropriate given the short duration and low workload of the task.

The first group consisted of \textbf{non-native readers}, represented by three undergraduate students with intermediate English proficiency. Each evaluator assessed a total of 80 randomly selected documents, comparing two summaries per document. They rated each summary independently on three criteria (coherence, simplicity, and faithfulness) using a five-point scale. In addition, they indicated their overall preference by selecting which of the two summaries they considered more readable and useful for a general audience.  

The second group consisted of \textbf{elementary school students}, represented by three participants with limited English proficiency (two fifth-grade students and one sixth-grade student).  As an ethical measure, these participants were assigned a filtered subset of summaries covering familiar or educationally accessible topics (e.g., general science or daily-life contexts), since the original materials were considered too complex for their age group. Given their age and reading level, the number of evaluation items was also intentionally limited to reduce fatigue and maintain engagement. Each evaluator reviewed 10 documents from this subset, providing ratings on the same three criteria and selecting their preferred summary in each case. For ethical reasons, we did not conduct evaluations with the \textbf{attention-deficit group}.  

To further assess annotator consistency, we computed Krippendorff’s $\alpha$ within each group. The results showed low agreement overall (generally ranging from –0.3 to 0.5), with relatively higher consistency only on the Simplification dimension for the elementary school student group ($\alpha$ = 0.52). Such low inter-annotator agreement is common in summarization tasks, especially with only three evaluators and inherently subjective criteria. The difference in language proficiency between non-native undergraduates and elementary-level readers likely contributed to variability in judgments. Given this heterogeneous evaluator composition and the subjective nature of the task, achieving high inter-rater agreement values, such as Krippendorff’s $\alpha$, is inherently challenging \cite{fabbri2021summeval}. Nevertheless, an analysis of the standard deviations across evaluators revealed that their ratings followed broadly similar tendencies, with average deviations of 0.40 for non-native readers and 0.54 for elementary school student readers (ranging from 0.30-0.69 across datasets), indicating a reasonable degree of consistency in practice. Overall, even if the numerical agreement metrics appear low, the evaluators largely showed similar judgments about which summaries were clearer and more faithful.

Figure~\ref{fig:human1} and Figure~\ref{fig:human2} show the evaluation platform used in the study, which offered a structured and user-friendly interface for the entire evaluation workflow, including both summary comparison and preference selection.

\begin{figure*}
    \centering
    \includegraphics[width=1.8\columnwidth]{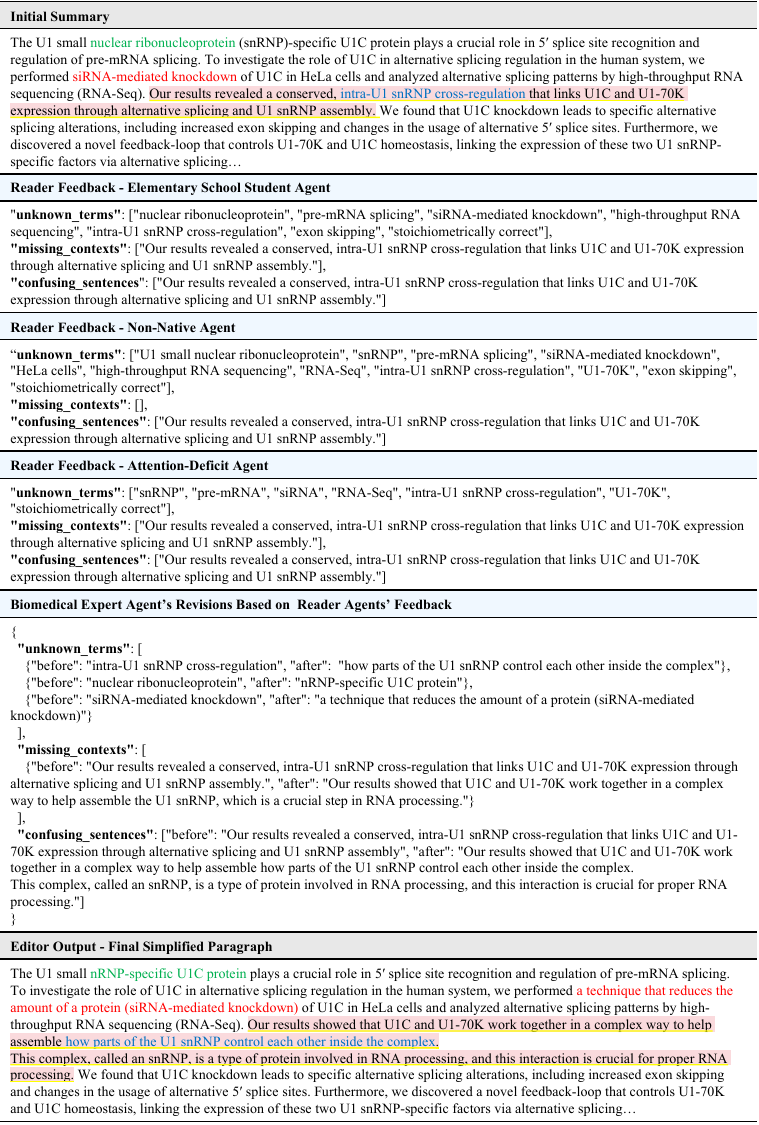}
    \caption{NRLB revision process for a biomedical paper. Reader agents identify readability barriers, the Domain Expert Agent proposes targeted edits, and the Editor Agent integrates them into a simplified summary. Colors, underlines, and highlights indicate unknown terms, missing contexts, and confusing sentences, respectively.}
    \label{fig:case_plos}
\end{figure*}

\begin{figure*}
    \centering
    \includegraphics[width=1.8\columnwidth]{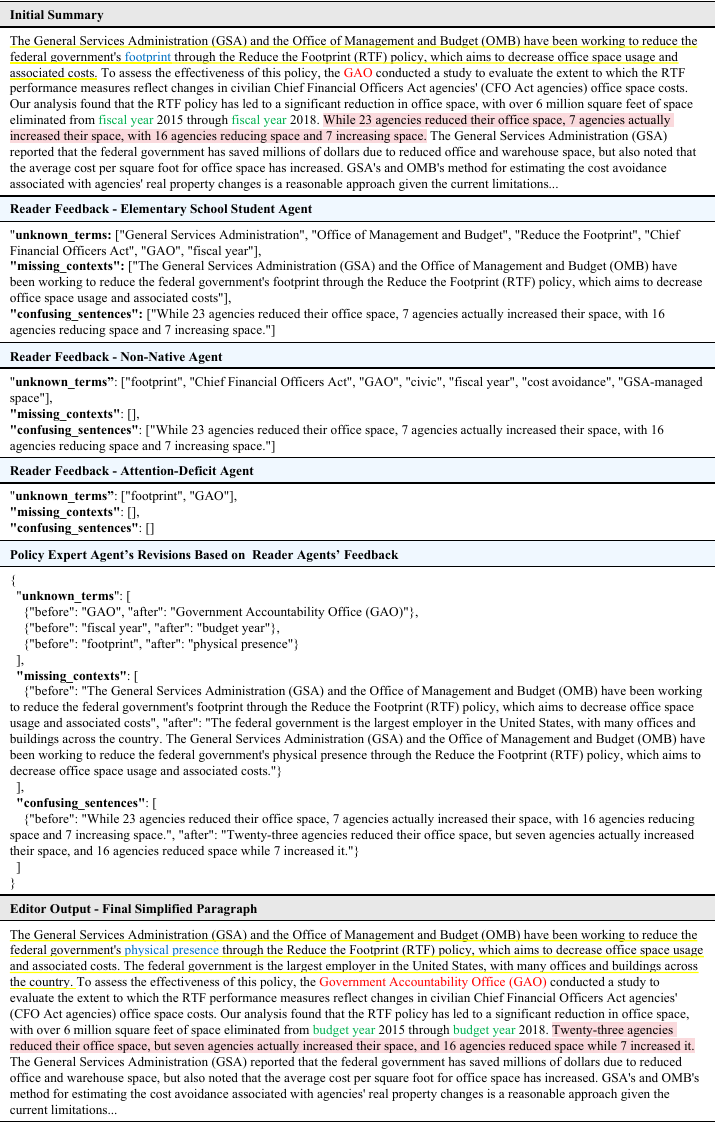}
    \caption{NRLB revision process for a policy document. Reader agents identify readability barriers, the Domain Expert Agent proposes targeted edits, and the Editor Agent integrates them into a simplified summary. Colors, underlines, and highlights indicate unknown terms, missing contexts, and confusing sentences, respectively.}
    \label{fig:case_gov}
\end{figure*}

\begin{figure*}
    \centering
    \includegraphics[width=1.8\columnwidth]{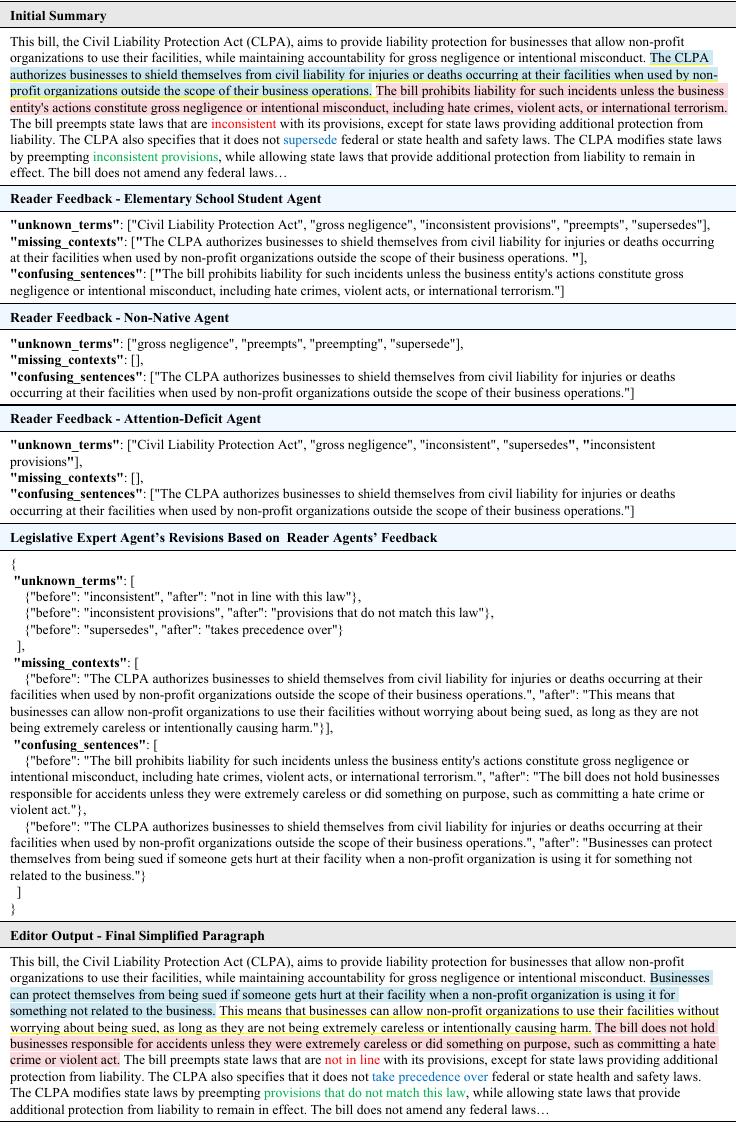}
    \caption{NRLB revision process for a legislative bill. Reader agents identify readability barriers, the Domain Expert Agent proposes targeted edits, and the Editor Agent integrates them into a simplified summary. Colors, underlines, and highlights indicate unknown terms, missing contexts, and confusing sentences, respectively.}
    \label{fig:case_bill}
\end{figure*}

\begin{figure*}
    \centering
    \includegraphics[width=1.8\columnwidth]{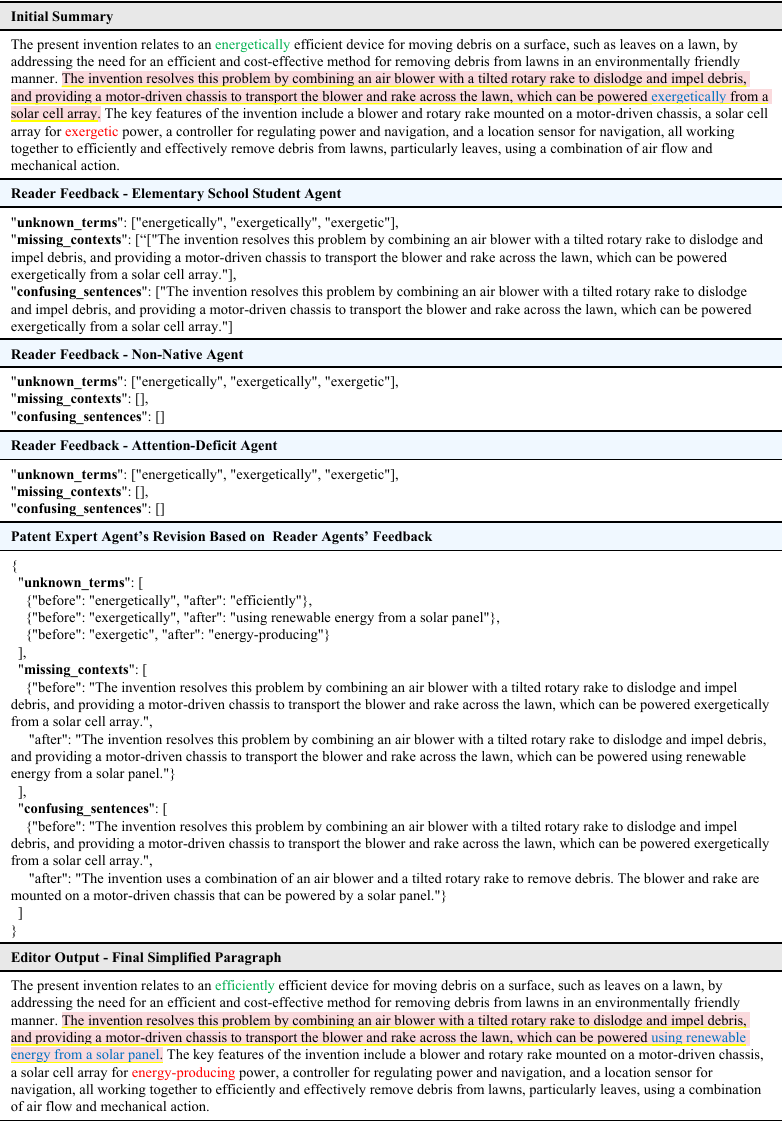}
    \caption{NRLB revision process for a patent document. Reader agents identify readability barriers, the Domain Expert Agent proposes targeted edits, and the Editor Agent integrates them into a simplified summary. Colors, underlines, and highlights indicate unknown terms, missing contexts, and confusing sentences, respectively.}
    \label{fig:case_big}
\end{figure*}

\section{Additional Case Studies}
\label{sec:F}
To illustrate how the NRLB framework operates across different document types, we present four representative case studies selected from the PLOS, GovReport, BillSum, and BigPatent datasets. Figures~\ref{fig:case_plos} through~\ref{fig:case_big} provide a step-by-step visualization of the full revision pipeline applied to one summary from each domain. These examples highlight how NRLB identifies and addresses readability barriers specific to each domain.

Each case begins with the generation of an initial summary, which is then evaluated by three agents representing an elementary school student reader, a non-native reader, and an attention-deficit reader. These agents independently annotate problematic expressions based on their simulated reading experiences. Feedback is categorized into unknown terms, missing contextual information, and confusing sentence structures. These issues often involve technical vocabulary, implicit references, or overly complex syntax that can hinder comprehension.

Based on this feedback, the Domain Expert Agent proposes targeted revisions in a clear before-and-after format. Suggested edits may involve replacing technical terms with simpler alternatives, adding brief explanations for unfamiliar entities, or restructuring convoluted sentences. Each edit is explicitly linked to a specific comprehension barrier identified by the reader agents, ensuring that the revisions address real accessibility challenges rather than arbitrary simplification.

The Editor Agent synthesizes the proposed edits into a revised summary that improves both clarity and factual consistency. The visualization highlights the original problem areas using colors for unknown terms, underlines for missing contexts, and highlights for confusing structures. Among the four domains, PLOS summaries required an additional revision round due to the density of biomedical terminology. In contrast, summaries from GovReport, BillSum, and BigPatent typically reached satisfactory readability within one or two rounds. These results demonstrate how the NRLB framework supports adaptive, agent-guided revision tailored to domain-specific complexity and reader needs.

\section{Failure Case Analysis}
\label{sec:G}
In this section, we analyze representative minor failure cases observed during the iterative simplification process. These cases are infrequent, structurally explainable within the multi-round framework, and can be partially mitigated through design choices (e.g., setting K = 3 and limiting revisions to Round 2).

\textbf{Failure Type 1}: Loss of Precision due to Over-Simplification
Aggressive simplification can reduce domain-specific precision when technical terms are replaced with general expressions without proper definition. The text remains broadly accurate, but legal, financial, or scientific nuances may be weakened. For example, generation logs show that “tax expenditures” was simplified to “tax breaks given to certain groups.” While this improves accessibility, it does not fully capture the formal definition referring to revenue losses through the tax system, similar to government spending. As this error mainly occurs in later stages, we limit revisions to Round 2 to prevent excessive generalization while maintaining readability.

\textbf{Failure Type 2}: Emergence of Latent Context Gaps during Simplification
Another failure arises when context-related issues not apparent earlier become visible after simplification, reflecting limited context awareness. Early stages focus on lexical difficulty, but as complexity decreases, gaps in background knowledge become more apparent. For instance, “The study of HIV-1 TAT-interacting protein (TIP60) has revealed its role as a tumor suppressor…” was initially flagged only for unknown terms, but after simplification, the Elementary School Student Reader Agent additionally identified missing background context. This shift from lexical to contextual difficulty is typical in iterative simplification and highlights the role of multi-round refinement in uncovering deeper comprehension barriers.

\end{document}